\documentclass[preprint,12pt,5p,twocolumn]{elsarticle}

\usepackage{amssymb}
\usepackage[utf8]{inputenc} 
\usepackage[T1]{fontenc} 
\usepackage[hidelinks]{hyperref} 
\usepackage{url} 
\usepackage{booktabs} 
\usepackage{amsfonts} 
\usepackage{amsmath}
\usepackage{amssymb}
\usepackage{nicefrac} 
\usepackage{microtype} 
\usepackage{doi}
\usepackage[caption=false, font=footnotesize]{subfig}
\usepackage[dvipsnames, table]{xcolor} 
\usepackage{multirow}
\usepackage{floatrow}
\usepackage{blindtext}
\usepackage{physics} 
\usepackage{bm} 
\usepackage{upgreek} 
\usepackage{listings} 
\usepackage[inline]{enumitem} 
\usepackage{dblfloatfix} 
\usepackage{tikz}
\usepackage{xcolor}
\usepackage{balance}

\makeatletter
\def\ps@pprintTitle{%
    \let\@oddhead\@empty
    \let\@evenhead\@empty
    \def\@oddfoot{\footnotesize\itshape
    {To appear in \@journal}\hfill\today}%
    \let\@evenfoot\@oddfoot
    }
\makeatother

\renewcommand{\bold}[1]{\mathbf #1}

\newcommand{\hidecomment}{} 
\newcommand{\showcomment}{}

\renewcommand{\showcomment}{Dummy text}

\newcommand{\comment}[1]{%
  \ifx\showcomment\hidecomment
  \else
    #1 
  \fi
}

\newcommand{\tempdone}[1] {}
\newcommand{\new}[1] {{\color{black} #1}}

\newcommand{\change}[1] {{\color{black} #1}}  

\newcommand{\logmap}{\text{Log}}
\newcommand{\expmap}{\text{Exp}}

\newcommand{\mysection} [1] {Sec.~\ref{#1}}
\newcommand{\myfigure} [1] {Fig.~\ref{#1}}
\newcommand{\myequation} [1] {Eq.~(\ref{#1})}
\newcommand{\mytable} [1] {Tab.~\ref{#1}}

\newcommand{\bo} [1] {\textbf{#1}}

\newfloatcommand{capbtabbox}{table}[][\FBwidth]

\tikzstyle{every path}=[line width=0.5pt]

\definecolor{mypink}{RGB}{255, 195, 195}
\definecolor{myblue}{RGB}{175, 199, 238}
\definecolor{myyellow}{RGB}{255, 229, 153}
\definecolor{mybrown}{RGB}{192, 157, 134}
\definecolor{mylightyellow}{RGB}{243, 243, 167}

\DeclareRobustCommand{\legendsquare}[1]{%
  \tikz[baseline=(a.south)]{\node[fill=#1, draw=black, inner sep=.8ex, outer sep=0] (a) {};}%
}

\journal{Robotics and Autonomous Systems}

\begin{document}

\begin{frontmatter}


\title{Continual Learning from Demonstration of Robotics Skills}

\date{}

\author[uibk]{Sayantan~Auddy\corref{cor1}}
\author[uibk]{Jakob~Hollenstein}
\author[trento]{Matteo~Saveriano}
\author[uibk]{\linebreak Antonio~Rodr\'{i}guez-S\'{a}nchez}
\author[uibk,disc]{Justus~Piater}

\cortext[cor1]{Corresponding author: \texttt{sayantan.auddy@uibk.ac.at}}
\address[uibk]{Department of Computer Science, University of Innsbruck, Austria.}
\address[trento]{Department of Industrial Engineering, University of Trento, Italy.}
\address[disc]{Digital Science Center (DiSC), University of Innsbruck, Austria.}
 
\begin{abstract}

Methods for teaching motion skills to robots focus on training for a single skill at a time. Robots capable of learning from demonstration can considerably benefit from the added ability to learn new movement skills without forgetting what was learned in the past. To this end, we propose an approach for continual learning from demonstration using hypernetworks and neural ordinary differential equation solvers. We empirically demonstrate the effectiveness of this approach in remembering long sequences of trajectory learning tasks without the need to store any data from past demonstrations. Our results show that hypernetworks outperform other state-of-the-art continual learning approaches for learning from demonstration. In our experiments, we use the popular LASA benchmark, and two new datasets of kinesthetic demonstrations collected with a real robot that we introduce in this paper called the \emph{HelloWorld} and \emph{RoboTasks} datasets. We evaluate our approach on a physical robot and demonstrate its effectiveness in learning real-world robotic tasks involving changing positions as well as orientations. We report both trajectory error metrics and continual learning metrics, and we propose two new continual learning metrics. Our code, along with the newly collected datasets, is available at \url{https://github.com/sayantanauddy/clfd}.

\end{abstract}




\end{frontmatter}


\begin{figure*}[t!]
	\includegraphics[width=\textwidth]{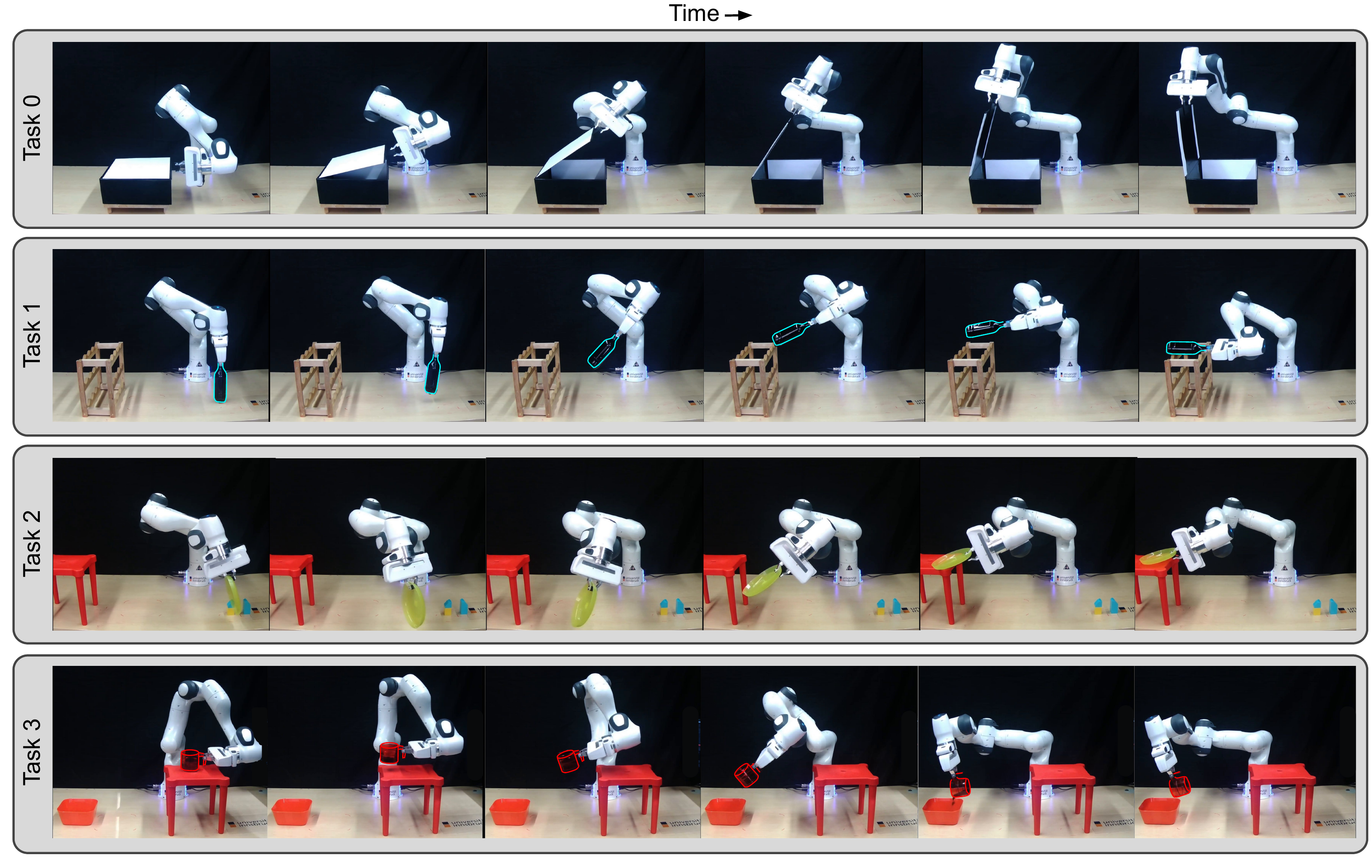}
	\caption{A robot, continually trained using learning from demonstration to perform \change{real-world} tasks involving changing positions and orientations of the end-effector, can reproduce all the tasks that it has learned in the past with negligible parameter growth and without retraining on past tasks. The order of tasks during sequential training is 0: \emph{box opening}, 1: \emph{bottle shelving}, 2: \emph{plate stacking}, 3: \emph{pouring}. Video of the robot performing these tasks is available at \url{https://youtu.be/0gdIImIBnXc}. Further details can be found in \mysection{sec:res_robtasks}.} 
	\label{fig:robottasks_prediction}
\end{figure*}


%
\section{Introduction}
\label{sec:intro}

Robots deployed in unstructured real-world environments will face new tasks and challenges over time, requiring capabilities that cannot be fully anticipated at the beginning. 
These robots need to learn continually, which implies that they should be able to acquire new capabilities without forgetting the previously learned ones. 
Furthermore, a continual learning robot should be able to do this without the need to store and retrain on the training data of all the previously learned skills.

Continual learning can be effective in expanding a robot's repertoire of skills and in increasing the ease of use for non-expert human users. However, apart from a few approaches for robotics \citep{gao2021cril, huang2021continual}, the current continual learning research mostly focuses on vision-based tasks such as incrementally learning classification of new image categories \citep{shin2017continual, aljundi2018MAS, von2019continual}.

Continually acquiring perceptive skills is important for a robot that interacts with its environment, but equally important is the ability to incrementally learn new movement skills. Learning from demonstration (LfD) \citep{billard2016learning} is a popular and tangible way to impart motion skills to robots, for instance via kinesthetic teaching, where a human user teaches new skills by guiding the robot. A trend that recently received increased attention in LfD is encoding observations into a vector field \citep{hersch2008dynamical, khansari2011learning, urain2020imitationflow, kolter2019learning, Ijspeert2002Movement, saveriano2021dynamic}. These methods, like many other works in the field, focus on learning a single motion. To naively learn multiple motion skills, one would need to train a different model for each skill, or jointly train on the demonstrations for all skills. 

In this paper, we propose an approach to continual learning from demonstration in which a robot learns individual motions sequentially 
\emph{without~retraining} on past demonstrations. The motion demonstrations are recorded in the robot's task space and consist of either trajectories of the end-effector position, or trajectories of the end-effector position and orientation (the robot is free to rotate in all 3 rotation axes). 
\new{
The skills learned from demonstrations of different tasks are incorporated into a common unified model, after which our robot can reproduce all the trajectories it has learned in the past (\myfigure{fig:robottasks_prediction}). To the best of our knowledge, \change{this is the first work on continual trajectory learning from demonstrations.
}
}

More specifically, we show that a Hypernetwork~\citep{ha2017hypernetworks, von2019continual}, that generates the parameters of a Neural Ordinary Differential Equation (NODE) solver~\citep{chen2018neural}, remembers a long sequence of motion skills equally well as when learning each task with a separate NODE. The number of parameters of the hypernetwork grows by a negligible amount for each new task, making it suitable for potential deployment on resource-constrained, non-networked robotic platforms. We also demonstrate the effectiveness of chunked hypernetworks~\citep{von2019continual} which can be even smaller in size than the NODEs they generate. Our results show how using a time index as an additional, direct input to a NODE increases its prediction accuracy for complex trajectories.
We first evaluate our approach on the popular LASA trajectory learning benchmark \citep{khansari2011learning}, where our model learns a long sequence of 26 tasks without forgetting.
Due to the lack of datasets containing real robot data useful for continual learning from demonstration, we introduce two new datasets, named \emph{HelloWorld} and \emph{RoboTasks}.
The HelloWorld dataset consists of two-dimensional trajectories of the robot's position collected with a Franka Emika Panda robot. 
RoboTasks is a dataset of 
\change{real-world}
robot tasks containing trajectories of the robot's position as well as its orientation in 3D space (\myfigure{fig:robottasks_prediction}). These two datasets serve as additional benchmarks to evaluate our approach, both quantitatively and qualitatively on a real robot.
For all three datasets, we extensively compare our hypernetwork-based continual LfD approach against methods from all major continual learning families (\emph{dynamic architecture}, \emph{replay}, and \emph{regularization}~\citep{parisi2019CL_survey}) and report multiple metrics, both for accuracy of the reproduced motion skills, as well as the continual learning performance.
\new{
Finally, we introduce 
two new easily-computable metrics that offer additional insights into the continual learning performance: \emph{Time Efficiency} (TE) and \emph{Final Model Size} (FS). 
TE captures training time changes while learning multiple tasks, and FS measures the relative model size after learning all tasks.
}

In summary, our primary contributions are:%
\begin{itemize}[noitemsep,topsep=0pt]
    \item We propose an approach to learning from demonstration with hypernetworks and NODEs for continually learning new tasks without reusing training data of previous tasks. We show that this approach can be used for learning robot tasks in the \change{real world}.
    \item We release 2 new datasets suitable for continual LfD: a dataset containing $7$ tasks of planar motions, and a dataset of 4 tasks of motions involving both position and orientation. Both datasets are collected with a real robot using kinesthetic teaching.  
\end{itemize}

%

\section{Related Work}
\label{sec:related}

\change{

Robot \emph{learning from demonstration} (LfD) is a means 
for humans to teach motion skills to robots without explicitly programming them \citep{billard2016learning}, 
which allows even users
without expertise in robotics to train robots. It is also known by other names such as \emph{programming by demonstration} or \emph{imitation learning} \citep{calinon2018learning, billard2016learning}. The demonstrations required for training robots via LfD can be obtained by different means, some of which are: 
\begin{enumerate*}[label=(\roman*)]
	\item using a motion-tracking system to record human motions,
	\item using teleoperation to operate the robot while recording the robot's state, or
	\item using kinesthetic teaching where a human user physically guides the robot to perform a motion task \citep{argall2009survey, billard2016learning, ravichandar2019robot, ahmadzadeh2018trajectory}.
\end{enumerate*} 
Once the demonstrations are available, there exist several different algorithmic approaches for learning from this data \citep{billard2016learning}. Supervised learning has been used to learn from either a single demonstration \citep{wu2010towards} or a collection of demonstrations \citep{argall2011teacher}. LfD has also been used in conjunction with reinforcement learning (RL) \citep{calinon2013compliant} where RL is used to refine the skills learned with LfD. Another approach is to learn a cost function from demonstrations and then to train a model predictive controller to reproduce the skills through inverse RL \citep{das2021model} or through constrained optimization \citep{englert2017inverse}. In addition to demonstrations which show the robot the motion it has to perform, negative demonstrations have also been shown to be advantageous \citep{kalinowska2021ergodic}. We refer the reader to \citep{argall2009survey, billard2016learning, ravichandar2019robot} for a comprehensive overview of LfD.

In this paper, we focus on a subfield of LfD: \emph{trajectory-based} learning methods that use a supervised approach for acquiring motion skills. These methods can be broadly categorized into two groups \citep{ahmadzadeh2018trajectory}:
}
some methods use \emph{generative models} to fit a distribution from the training data ~\citep{hersch2008dynamical, khansari2011learning, urain2020imitationflow}, while other methods
exploit \emph{function approximators} like neural networks to fit the training data~\citep{kolter2019learning, Ijspeert2002Movement}.
In both groups, training data can be used to learn a \emph{static mapping} (time~input $\rightarrow$ desired position) or a \emph{dynamic mapping} (input~position $\rightarrow$ desired velocity). A dynamic mapping generates vector fields where input quantities are transformed into their time derivatives, and different strategies have been proposed to ensure convergence of the vector field to a given target~\citep{kolter2019learning, khansari2014learning, saveriano2020energy, urain2020imitationflow}. 
\new{
Among others, the \emph{Imitation Flow} (iFlow) approach of Urain et al. \cite{urain2020imitationflow} leverages the representational power of neural networks and normalizing flows to learn vector fields from demonstrations. Another neural network based approach for learning vector fields is \emph{Neural Ordinary Differential Equation solvers} (NODEs) \citep{chen2018neural}. 
Although NODEs have not been exploited for LfD, in our experiments we found that their empirical performance is comparable to that of iFlow. Crucially, the time needed to train a NODE to convergence is significantly less than that required for an iFlow model with an equivalent parameter size. 
}

\change{Trajectory-based} LfD is a mature research field, but most methods assume that different tasks are encoded in different representations, i.e., one has to fit a new model for each task the robot has to execute \change{\citep{billard2016learning}}. In this paper, we take the continual learning perspective on learning by demonstration and propose an approach capable of continuously learning new tasks without needing to store and use the training data from past tasks.    

\smallskip

\noindent\emph{Continual Learning} approaches in the current literature mostly address the problem of continual image classification. 
Popular strategies include growing the network architecture \citep{rusu2016progressive}, replaying data from past tasks, or regularizing trainable parameters to avoid catastrophic forgetting~\citep{parisi2019CL_survey}.
\emph{Replay}-based methods cache samples of real data from past tasks \citep{rebuffi2017icarl}, or use generative models to create pseudo-samples of past data \citep{shin2017continual}, which are interleaved with the current task's data during training. 
\emph{Regularization}-based methods~\citep{zenke2017SI, aljundi2018MAS} add a regularization term to the learning objective to minimize changes to parameters important for solving previous tasks. 
Refer to \citep{parisi2019CL_survey, delange2021continual} for in-depth surveys on continual learning methods.

Continual learning has also been successfully applied to robotics, though the number of such studies is relatively few.
Gao et al.~\cite{gao2021cril} present an approach for continual imitation learning that relies on deep generative replay \citep{shin2017continual} and action-conditioned video prediction to generate state and action trajectories of past tasks. This pseudo-data is interleaved with demonstrations of the current task to train a policy network that controls the robot's actions.
The authors note that the generation of high-quality video frames can be problematic for a long sequence of tasks. 

Xie and Finn~\cite{xie2021lifelong} propose a method for lifelong robotic reinforcement learning that seeks to improve the forward transfer performance while learning a new current task by pre-training on the entire experience collected from all previous tasks. The problem of catastrophic forgetting is not considered.

Our continual LfD approach shares some similarities with the work of Huang et al.~\cite{huang2021continual} who continually learn a dynamics model for reinforcement learning.
In their work, a task-conditioned hypernetwork generates the parameters of the dynamics model for reinforcement learning tasks such as opening doors or pushing blocks. In contrast, we use hypernetworks for generating parameters for a trajectory learning NODE in a setup for learning from demonstration. We follow a supervised approach and do not need to rely on robot simulators. %
Compared to \cite{huang2021continual}, we evaluate on much longer sequences of tasks and also investigate the effectiveness of chunked hypernetworks \citep{von2019continual}. In addition, we qualitatively evaluate our approach on a physical robot.

\new{
LfD involves learning the entire vector field of a robot's motion from only a handful of demonstration trajectories.
This makes it a more challenging problem than typical supervised regression or classification scenarios where the amount of training data is usually much greater.
To continually learn such vector fields of different kinds of motion demonstrations, we need to tackle the challenges of LfD as well as the problem of \emph{catastrophic forgetting} associated with neural networks. 
As far as we know, ours is the first work that demonstrates that continual LfD is feasible for sequences of 
\change{real-world}
robot tasks.
}


\section{Background}
\label{sec:background}

In this paper, we utilize Neural Ordinary Differential Equation (NODE) solvers~\citep{chen2018neural} to learn from demonstrations and different state-of-the-art continual learning approaches \citep{zenke2017SI, aljundi2018MAS, von2019continual} to alleviate catastrophic forgetting~\citep{parisi2019CL_survey} when the NODEs learn a sequence of multiple tasks. 

\subsection{Learning from Demonstration}
\label{sec:back_lfd}

In the simplest case, task space demonstrations provided to a robot can consist of trajectories of only the end-effector positions. Since these position trajectories reside in Euclidean space, we can directly use a NODE to learn them using maximum likelihood estimation. However, the more general situation is when demonstrations contain both the position and orientation (which are commonly expressed in terms of unit quaternions). Trajectories of unit quaternions do not reside in Euclidean space and require additional processing, as discussed in Sec.~\ref{sec:back_ori_traj}. 

\subsubsection{Neural ODE Solver}
\label{sec:back_node}

Consider a set of $N$ observed trajectories $\mathcal{D}=\{\bold{y}^{(0)}_{0:T-1}, \ldots, \bold{y}^{(N-1)}_{0:T-1}\}$, where each trajectory $\bold{y}^{(i)}_{0:T-1}$ is a sequence of $T$ observations $\bold{y}^{(i)}_{t} \in \mathbb{R}^d$. 
Each observation $\bold{y}^{(i)}_{t}$ is a perturbation of an unknown true state $\bold{x}^{(i)}_{t}$ generated by an unknown underlying vector field $\bold{f}_{\text{true}}$ \citep{heinonen2018learning}:
\begin{equation}
\bold{x}_t = \bold{x}_0 + \int_{0}^{t}\bold{f}_{\text{true}}(\bold{x}_{\tau})\ \dd\tau ,
\label{eq:node_integral}
\end{equation}
where $\bold{x}_0$ is the true starting state of the trajectory. The goal of a Neural Ordinary Differential Equation (NODE) solver \citep{chen2018neural} is to learn a neural network $\bold{f}_{\bm{\uptheta}}$ parameterized by $\bm{\uptheta}$ that approximates
the true underlying dynamics of the observed system such that $\bold{f}_{\boldsymbol\uptheta} \approx \bold{f}_{\text{true}}$.
As we do not have access to $\bold{f}_{\text{true}}$ but only to the noisy observed trajectories, we compute the loss $\mathcal{L}$ based on the difference of the forward simulated states of the NODE $\hat{\bold{y}}_{t}$ and the observations $\bold{y}_{t}$:

\begin{align}
&\mathcal{L} = \cfrac{1}{2} \sum_{t} \parallel\bold{y}_{t} - \hat{\bold{y}}_{t}\parallel^2_2 \label{eq:node_loss}\\
&\text{ where } \hat{\bold{y}}_t = \hat{\bold{y}}_0 + \int_{0}^{t}\bold{f}_{\bm\uptheta}(\hat{\bold{y}}_{\tau})\,\ \dd\tau \nonumber
\end{align}
When the set of observed trajectories $\mathcal{D}$ consists of only positions, then the demonstrations can be learned directly with the help of \myequation{eq:node_loss}.

\subsubsection{Learning Orientation Trajectories}
\label{sec:back_ori_traj}
When demonstration trajectories of the robot's orientation are expressed as unit quaternions, it is not possible to directly utilize \myequation{eq:node_loss} due to the unit norm constraint \citep{huang2020toward}. Following Ude et al.~\cite{ude2014orientation} and Huang et al.~\cite{huang2020toward}, we project quaternions into the tangent space which we can consider a local Euclidean space. These transformed trajectories can then be learned with \myequation{eq:node_loss}. For inference, the Euclidean trajectories predicted by the NODE are transformed back into quaternions and then passed to the robot.

Consider a set of $N$ observed orientation trajectories $\mathcal{D}_q=\{\bold{q}^{(0)}_{0:T-1}, \ldots, \bold{q}^{(N-1)}_{0:T-1}\}$, where the $i^\text{th}$ trajectory $\bold{q}^{(i)}_{0:T-1}$ is a sequence of $T$ quaternions $\bold{q}^{(i)}_{t} = \begin{bmatrix}
								           v_{t}^{(i)} \\
								           \mathbf{u}_{t}^{(i)} 
								         \end{bmatrix}$, where $\bold{q}^{(i)}_{t} \in \mathbb{S}^3$,  $v_{t}^{(i)} \in \mathbb{R}$ and $\mathbf{u}_{t}^{(i)} \in \mathbb{R}^3$. We convert $\bold{q}^{(i)}_{t}$ into a rotation vector $\bold{r}^{(i)}_{t} \in \mathbb{R}^3$ using 
\begin{equation}
	\mathbf{\bold{r}^{(i)}_{t}} = \logmap(\bar{\bold{q}}^{(i)}_{t} * \bold{q}^{(i)}_{T-1}) 
\label{eq:logmap1}
\end{equation}
where $\bar{\bold{q}}$ indicates the conjugate of the quaternion $\bold{q}$ and the $*$ operator denotes quaternion multiplication,
and $\logmap(\cdot): \mathbb{S}^3 \mapsto \mathbb{R}^3$ is the \emph{logarithmic map} 
\citep{saveriano2019merging}, defined as
\begin{equation}
	\logmap(\mathbf{q}) = \begin{cases}
	            		 \arccos(v)\frac{\mathbf{u}}{||\mathbf{u}||} \text{ if } ||\mathbf{u}|| > 0, \\
						 [0, 0, 0]^\text{T} \text{ otherwise.}
						 \end{cases}
\label{eq:logmap2}
\end{equation}
Here, $\bold{q}^{(i)}_{T-1}$ is the final quaternion in the sequence $\bold{q}^{(i)}_{0:T-1}$ and $\bar{\bold{q}}^{(i)}_{t}$ is the conjugate of $\bold{q}^{(i)}_{t}$. By applying \myequation{eq:logmap1} on the trajectories in $\mathcal{D}_q$, we obtain a set of Euclidean trajectories $\mathcal{D}_r$, which can then be learned using \myequation{eq:node_loss}. After learning is complete, the trajectories $\hat{\mathcal{D}}_r=\{\hat{\bold{r}}^{(0)}_{0:T-1}, \ldots, \hat{\bold{r}}^{(N-1)}_{0:T-1}\}$ predicted by the NODE can be converted into quaternion trajectories $\hat{\mathcal{D}}_q$ using
\begin{equation}
	\mathbf{\hat{\bold{q}}^{(i)}_{t}} = \bold{q}^{(i)}_{T-1} * \expmap(\hat{\bold{r}}^{(i)}_{t}) 
\label{eq:expmap1}
\end{equation}
where $\expmap(\cdot): \mathbb{R}^3 \mapsto \mathbb{S}^3$ is the \emph{exponential map} 
\citep{saveriano2019merging}, defined as
\begin{equation}
	\expmap(\mathbf{r}) = \begin{cases}
						 \change{
						 \left[\cos(||\mathbf{r}||), \sin(||\mathbf{r}||)\frac{\mathbf{r}^\text{T}}{||\mathbf{r}||}\right]^\text{T}} \text{ if } ||\mathbf{r}||>0, \\
						 \left[1, 0, 0, 0\right]^\text{T} \text{ otherwise.}
						 \end{cases}
\label{eq:expmap2}
\end{equation}
Once we obtain $\hat{\mathcal{D}}_q$, it can be directly compared against the ground truth demonstrations $\mathcal{D}_q$. However, we assume that the input domain of $\logmap(\cdot)$ is restricted to $\mathbb{S}^3$ except for $\left[1, 0, 0, 0\right]^\text{T}$ and the input domain of $\expmap(\zeta)$ is constrained to satisfy $||\zeta||<\pi$ \citep{huang2020toward}.


\subsection{Continual Learning}
\label{sec:back_cl}

Existing continual learning approaches can be broadly categorized into a few groups, the most prominent among which are methods based on \emph{dynamic architectures}, methods based on \emph{replaying} (or \emph{pseudo-rehearsing} on) training data of past tasks, and methods based on \emph{regularization} \citep{parisi2019CL_survey}. 
\change{
In this paper, we consider continual learning methods (CL methods) from all of these categories.  
}

\subsubsection{Dynamic Architectures}
\label{sec:back_cl_dyna_arch}

\noindent\bo{Progressive Networks}: One of the early continual learning approaches, proposed by Rusu et al.~\cite{rusu2016progressive}, involves the addition of a new network for each new task, while reusing feature-mapping knowledge from previous tasks through lateral layer-wise connections from the networks of previous tasks. Although this approach eliminates catastrophic forgetting by design, it does not scale well to a large number of tasks due to the unconstrained growth of parameters and it also has the problem of a gradual slowing down of inference due to the increasing number of lateral connections. 

\subsubsection{Replay}
\label{sec:back_cl_rep}

Some continual learning approaches are based on the idea of replaying the training data (or some part of the data, or a compressed version of the data) of previous tasks while learning a new task \citep{rebuffi2017icarl}. While learning task $k$, the most naive way to do this is to combine the data from older tasks $\{\mathcal{D}_0, \cdots, \mathcal{D}_{k-1}\}$ with the data of the current task $\mathcal{D}_k$ to get a combined dataset $\{\mathcal{D}_0, \cdots, \mathcal{D}_{k}\}$. Thus, the network has access to all the training data from every task. However, this approach is not scalable due to the linear storage requirements with the number of tasks.

\subsubsection{Regularization}
\label{sec:back_cl_reg}

\noindent\bo{Synaptic Intelligence}: Synaptic Intelligence (SI) \citep{zenke2017SI} is a regularization-based continual learning approach. Each neural network parameter is assigned an importance measure based on its contribution to the change in the loss.
The loss for the $m^\text{th}$ task is defined as:
\begin{equation}
\tilde{\mathcal{L}}^m = \mathcal{L}^m + c \sum_{k} \Omega^m_k \left( \theta_k^* - \theta_k\right)^2 ,
\label{eq:si_loss}
\end{equation}
where $c$ is the regularization constant which trades off between learning a new task and remembering previously learned tasks, $\theta_k^*$ denotes the value of the $k^{\text{th}}$ parameter before starting to learn the $m^\text{th}$ task, and $\theta_k$ is the current value of the $k^{\text{th}}$ parameter. The per-parameter regularization strength $\Omega^m_k$~\citep{zenke2017SI} is given by
\begin{equation}
\Omega^m_k = \sum_{l<m} \cfrac{\omega_k^l}{(\Delta^l_k)^2 + \xi} \, ,
\label{eq:si_regu}
\end{equation}
where $\omega^l_k$ is the importance of parameter $k$ for learning task $l$, $\Delta^l_k$ is the change in parameter $k$, and $\xi$ is a damping constant.
\smallskip

\noindent\bo{Memory Aware Synapses}: Memory Aware Synapses (MAS) \citep{aljundi2018MAS} is also a regularization-based continual learning approach. The loss for the $m^\text{th}$ task for MAS has the same form as SI (\myequation{eq:si_loss}). MAS differs from SI in the way $\Omega^m_k$ is computed: the importance of a trainable parameter depends on the 
gradient of the squared L$_2$ norm of the network's output, i.e.,
\begin{equation}
\Omega^m_k = \cfrac{1}{N} \sum_{n=1}^{N} \vert\vert g_k(\bold{x}_n) \vert\vert = \cfrac{1}{N} \sum_{n=1}^{N} \biggl\vert\biggl\vert \cfrac{\partial \text{L}_2^2 \left(\bold{f}_{\bm\uptheta} \left(\bold{x}_n \right) \right)}{\partial \theta_k} \biggl\vert\biggl\vert .
\label{eq:mas_regu}
\end{equation}
The above summation is performed over $N$ input data points.
\smallskip

\noindent\bo{Hypernetworks}:
A hypernetwork \citep{ha2017hypernetworks, von2019continual} is a meta-model that generates the parameters of a target network that solves the task we are interested in. It uses a trainable task embedding vector as an input to generate the network parameters for a task. Though the parameters $\bold{h}$ of the hypernetwork $\bold{f}_{\bold{h}}$ are regularized, the parameters $\bm\uptheta^{m+1}$ produced by a hypernetwork for the $(m+1)^\text{th}$ task can be arbitrarily far away in parameter space from the parameters $\bm\uptheta^{m}$ produced for the previous $m^\text{th}$ task. 
Intuitively, this gives a hypernetwork more freedom to find good solutions for both the $m^\text{th}$ and $(m+1)^\text{th}$ tasks than other regularization-based approaches~\citep{zenke2017SI, aljundi2018MAS}.

A two-step optimization process is used for training a hypernetwork~\citep{von2019continual}. First, a candidate change $\Delta\bold{h}$ for the hypernetwork parameters is computed which minimizes the task-specific loss $\mathcal{L}^m$ for the (current) $m^\text{th}$ task with respect to $\bm\uptheta^m$:
\begin{align}
&\mathcal{L}^m = \mathcal{L}^m(\bm\uptheta^m, \bold{y}^m) \text{ where } \bm\uptheta^m = \bold{f}_\bold{h}(\bold{e}^m, \bold{h})
\label{eq:hn_loss_1}
\end{align}
Here $\mathbf{e}^m$ is the task embedding vector and $\mathbf{y}^m$ is the data for the $m^\text{th}$ task.
Next, $\Delta\bold{h}$ is considered to be fixed and the actual change for the hypernetwork parameters $\bold{h}$ is learned~\citep{von2019continual} by minimizing the regularized loss $\tilde{\mathcal{L}}^m$ with respect to $\bm\uptheta^m=\bold{f}_\bold{h}(\bold{e}^m, \bold{h})$:
\begin{align}
\tilde{\mathcal{L}}^m &= \mathcal{L}^m\big(\bm\uptheta^m, \bold{y}^m\big) \nonumber \\
 &\quad + \cfrac{\beta}{m-1} \sum_{l=0}^{m-1} \big\vert\big\vert \bold{f}_{\bold{h}}(\bold{e}^l, \bold{h}^*) - \bold{f}_{\bold{h}}(\bold{e}^l, \bold{h} + \Delta\bold{h})\big\vert\big\vert^2
\label{eq:hn_loss_2}
\end{align}
Here $\bold{h}^*$ denotes the hypernetwork parameters before learning the $m^\text{th}$ task, and $\beta$ is a hyperparameter that controls the regularization strength. To calculate the second part of \myequation{eq:hn_loss_2}, the stored task embedding vectors $\{\bold{e}^0, \bold{e}^1, \dots, \bold{e}^l, \dots, \bold{e}^{m-1}\}$ for all tasks before the $m^\text{th}$ task are used. 
In each learning step, the current task embedding vector $\bold{e}^m$ is also updated to minimize the task-specific loss $\mathcal{L}^m$~\citep{von2019continual}.
Note that the parameters of the target network $\bm\uptheta^m$ for the $m^\text{th}$ task are simply the hypernetwork outputs and are not directly trainable.

Chunked hypernetworks~\citep{von2019continual} produce the parameters of the target network in segments known as \textit{chunks}. A regular hypernetwork has a very high-dimensional output, but a chunked hypernetwork's output is of a much smaller dimension, leading to a lower hypernetwork parameter size.
A chunked hypernetwork requires additional inputs in the form of trainable \emph{chunk embedding vectors}. While each task has its dedicated task embedding vector, chunk embedding vectors are shared across tasks and are regularized in the same way as the hypernetwork parameters. The task embedding vector and all the chunk embedding vectors are combined in a batch and fed into the hypernetwork to produce the target network parameters for a task in one forward pass~\citep{von2019continual}.

%

\section{Methods}
\label{sec:methods}

\begin{figure}[b!]
\centering
\includegraphics[width=\textwidth]{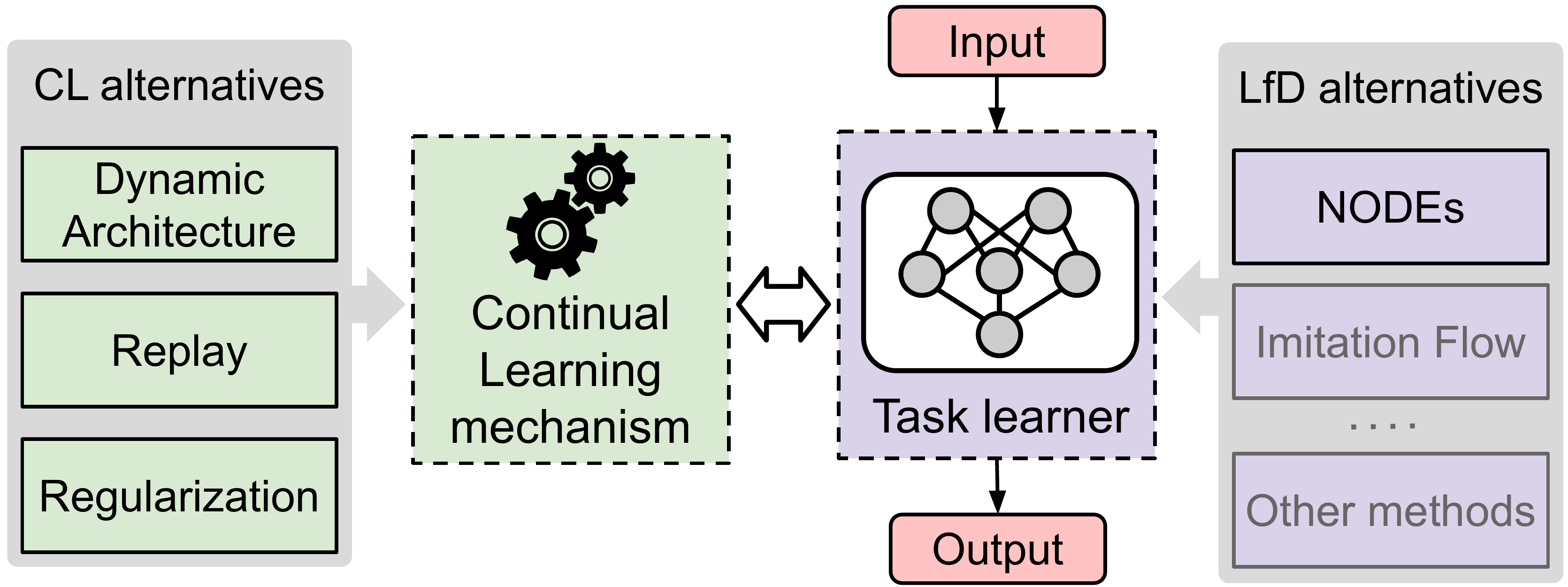}
\caption{
\new{
Overview of a continual learning system. The task learner performs the task we are interested in (purple dashed box) and the continual learning mechanism (green dashed box) mitigates the effect of catastrophic forgetting. The continual learning (CL) mechanism can be one of many alternatives: dynamic architecture-based, replay-based, or regularization-based. For learning from demonstration~(LfD), the task learner can be any trajectory learning approach, but we use NODEs for LfD (reasons are detailed in \mysection{sec:met_lfd}). Methods depicted with gray text are not considered for continual LfD in this paper.
} 
}
\label{fig:arch_high}
\end{figure}

As depicted in \myfigure{fig:arch_high}, any continual learning system in general consists of 2 components: 
\begin{enumerate*}[label=(\roman*)]
  \item a task learner, and
  \item a continual learning mechanism.
\end{enumerate*}
The task learner performs the task we are interested in; e.g.\ for continual learning of image classification, the task learner can be a CNN classifier. In this paper the task that we are interested in is \emph{learning from demonstration} (LfD), and the task learner in our case is a NODE as described in \mysection{sec:back_node}. 
The continual learning mechanism mitigates the effects of catastrophic forgetting as the task learner is trained on a sequence of tasks. This mechanism can act directly on the parameters of the task learner and can also contain other sub-components to achieve this goal. 	
\new{In this section, we describe in detail our proposed approach for continual learning from demonstration, as well as the other models used in our experiments. 
We also discuss the design choices made in this paper regarding the task learner and the continual learning mechanism.}

\subsection{Learning from Demonstration}
\label{sec:met_lfd}

Along with the basic NODE $\bold{f}_{\bm\uptheta}(\hat{\bold{y}}_t)$ described in \mysection{sec:back_node}, we use another variant where the NODE neural network $\bold{f}_{\bm\uptheta}(\hat{\bold{y}}_t, t)$ is a function of \change{both the state $\hat{\bold{y}}_t$ and the normalized time $t$ of the state ($t$ scales linearly from $t$=0.0 for the starting state to $t=N/f$ for the $N^{\text{th}}$ state of the trajectory, where $f$ is the recording frequency)}. This explicit time input results in the NODE learning a time-evolving vector field. We show empirically that this improves the accuracy of predicted trajectories, especially for those containing loops. We refer to this time-dependent NODE as NODE$^{\text{T}}$, and to the time-independent one as NODE$^{\text{I}}$. 

\new{
There exist many approaches for LfD \citep{urain2020imitationflow, khansari2014learning}, and in principle, any such method can be used as the \emph{task learner} (in \myfigure{fig:arch_high}).
In this paper we focus on neural network-based continual learning. Hence methods such as \emph{Stable Estimator of Dynamical Systems} (SEDS) \citep{khansari2014learning} are not considered as the constrained parameters (covariance matrices in case of SEDS) required by such methods cannot be produced by a neural network in a straightforward way. 
Alternative options for the task learner can be deep learning methods such as \emph{Imitation Flow} (iFlow) \citep{urain2020imitationflow} or NODEs \citep{chen2018neural}. We performed an experimental comparison between NODEs and iFLow (details in \mysection{sec:iflow} in the appendix) and found that the empirical performance of NODEs is better than iFlow and their training converges much quicker. Consequently, we use NODEs as the task learner in all the models used in our experiments.
}

\begin{figure*}[t!]
\centering
\includegraphics[width=\textwidth]{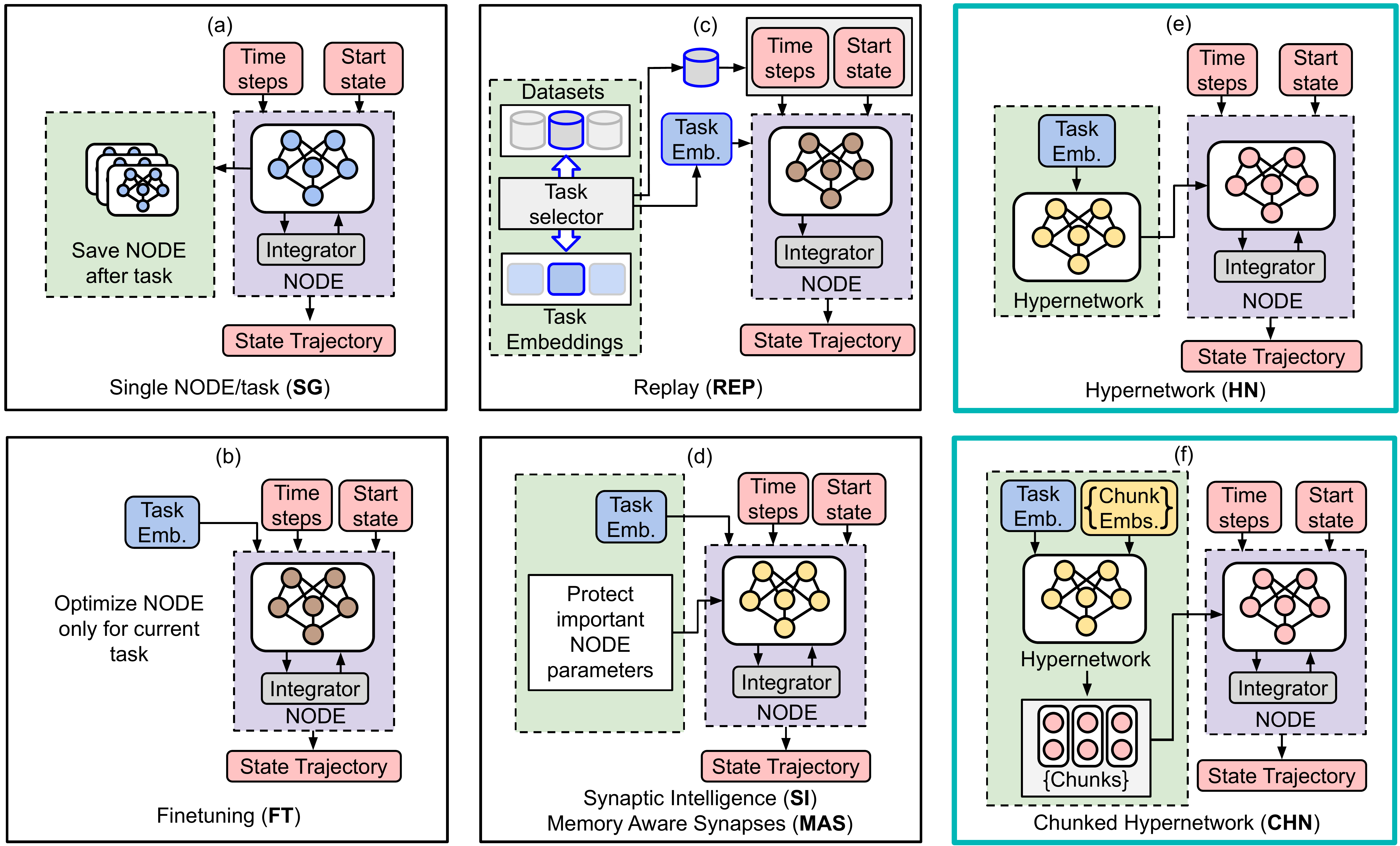}
\caption{Continual learning models used in our experiments are depicted in (a)-(f). Non-regularized trainable parameters that are initialized before and saved after each task are shown with \legendsquare{myblue}. Non-regularized parameters that are initialized only once before learning the first task are shown with \legendsquare{mybrown}. Regularized trainable parameters that are protected from catastrophic forgetting while learning a sequence of tasks are shown with \legendsquare{myyellow}. Other inputs and outputs are not trainable (shown with \legendsquare{mypink}). Given a start state and time steps, a NODE generates the state trajectory for the time steps.
(a)~\textbf{Single NODE per task~(SG)}: A separate NODE learns only a single task. This forms the upper-performance baseline. 
(b)~\textbf{Finetuning~(FT)}: The NODE from the previous task is optimized for the current task by finetuning.
(c)~\textbf{Replay~(REP)}: The training data from previous tasks is combined with the data of the current task.
(d)~\textbf{Synaptic~Intelligence~(SI)} and Memory Aware Synapses (\textbf{MAS}): NODE parameters are regularized to prevent changes to parameters important for solving previous tasks.
(e)~\textbf{Hypernetworks~(HN)}: A hypernetwork produces all the NODE parameters using a task embedding vector.
(f)~\textbf{Chunked~Hypernetworks~(CHN)}: Chunk embedding vectors together with a task embedding vector are used to produce the NODE parameters in segments called chunks. 
HN and CHN (highlighted in turquoise) are our proposed solutions for continual learning from demonstration.
}
\label{fig:arch_low}
\end{figure*}

\subsection{Baselines and Continual Learning Models}
\label{sec:met_cont_learning}

We enable continual learning for NODEs by adapting several state-of-the-art continual learning approaches for LfD. We consider all major families of continual learning (dynamic architecture, replay, and regularization) and describe the details of \change{each continual learning method (CL method)} used in our experiments (\myfigure{fig:arch_low}). 

\subsubsection{Dynamic Architecture}
\label{sec:met_dyna_arch}

\par{\noindent\bo{Single NODE per task (SG)}:
A simple way to learn $M$ tasks is to use a dedicated, newly-initialized NODE to learn a task and to freeze it afterwards. At the end we get $M$ NODEs, from which we can pick one at prediction time to reproduce the desired trajectory (\myfigure{fig:arch_low}(a)). In this setting, which acts as an upper-performance baseline, catastrophic forgetting is eliminated because the parameters of a NODE trained on a task are not affected when a new NODE is trained on the next task. However, this also means that we end up with $M$ times the number of parameters of a single NODE. 
This setup acts as a simplified version of the \emph{Progressive Networks} approach \citep{rusu2016progressive}, but without the layer-wise lateral connections to the networks of previous tasks. This keeps the inference time constant as new tasks are added.
}

\subsubsection{Optimizing only for the Current Task}
\label{sec:met_ft}

\par{\noindent\bo{Finetuning (FT)}: A single NODE is sequentially finetuned on $M$ tasks. To tell the NODE which task it should reproduce, i.e. to make it task-conditioned, we use an additional input in the form of a trainable vector known as a task embedding vector. This is similar to the approach followed for hypernetworks.
After the $m^\text{th}$ task is learned, the trained task embedding vector $\mathbf{e}^m$ for that task is saved. To reproduce the trajectory for the $m^\text{th}$ task, we pick the corresponding task embedding vector and use it as the additional network input. The NODE parameters are finetuned to minimize the loss on the current task without any mechanism for avoiding catastrophic forgetting. In this setting (\myfigure{fig:arch_low}(b)), we would expect the NODE to only remember the latest task and so this acts as a lower-performance baseline.
}

\subsubsection{Replaying Training Data}
\label{sec:met_replay}

\par{\noindent\bo{Replay (REP)}: For most continual learning scenarios \emph{replaying} training data from previous tasks is a trivial exercise of combining the data from disparate tasks and then training a model with the composite dataset. However, for learning from demonstration, such a simple approach does not suffice as our NODE models need to be task conditioned. If they are trained using a jumbled up combination of demonstration trajectories from different tasks, they will be unable to learn any of the tasks. To overcome this issue we propose to use a task selector, as shown in \myfigure{fig:arch_low}(c). REP stores the dataset of each task it learns, and similar to FT, also maintains separate task embeddings for each task. In each training iteration during learning task $m$, the task selector randomly selects 1 out of the $m$ previous tasks, and then passes the selected dataset and task embedding to the NODE, which optimizes its parameters for the selected task in that training iteration. When the REP model is trained for a moderately high number of iterations, each past task is sampled approximately uniformly and the NODE is able to learn continually. For learning new tasks, instead of scaling the number of iterations linearly with the number of tasks, we use a fixed number of training iterations, irrespective of the task being learned so that the run time does not explode while learning a long sequence of tasks.
}

\subsubsection{Regularization}
\label{sec:met_regu}

\par{\noindent\bo{Synaptic Intelligence (\textbf{SI})}: To learn $M$ tasks, the NODE parameters are regularized with the SI loss $\tilde{\mathcal{L}}^m$~(\myequation{eq:si_loss}).
The task-specific part ($\mathcal{L}^m$) of $\tilde{\mathcal{L}}^m$ corresponds to the NODE loss~(\myequation{eq:node_loss}).
Similar to FT, we make the SI NODE task-conditioned using a trainable task embedding vector, as shown in \myfigure{fig:arch_low}(d).
}
\smallskip

\par{\noindent\bo{Memory Aware Synapses (MAS)}: For MAS \citep{aljundi2018MAS}, we also follow the architecture in \myfigure {fig:arch_low}(d). The NODE parameters are learned using \myequation{eq:si_loss} and we use \myequation{eq:mas_regu} to compute $\Omega^m_k$. As before, we apply \myequation{eq:node_loss} as the task-specific loss $\mathcal{L}^m$. The MAS NODE is also made task-conditioned with a trainable task embedding vector.
}

Both SI and MAS seek to change the plasticity/elasticity of the parameters of the NODE in an attempt to prevent changes to parameters that are important for solving past tasks.
\smallskip

\par{\noindent\bo{Hypernetworks (HN)}: We use a hypernetwork to generate the parameters of a NODE.
We first compute the candidate change $\Delta\bold{h}$ for the hypernetwork parameters by minimizing the NODE loss (\myequation{eq:node_loss}), which acts as our task-specific loss $\mathcal{L}^m$:
\begin{align}
&\mathcal{L}^m = \mathcal{L}^m(\bm\uptheta^m, \bold{y}^m) = \cfrac{1}{2} \sum_{t} \parallel\bold{y}^m_{t} - \hat{\bold{y}}^m_{t}\parallel^2_2 \label{eq:our_hn_loss_1} \\
&\text{ where } \bm\uptheta^m = \bold{f}_\bold{h}(\bold{e}^m, \bold{h}),  \hat{\bold{y}}^m_t = \hat{\bold{y}}^m_0 + \int_{0}^{t}\bold{f}_{\bm\uptheta^m}(\hat{\bold{y}}_{\tau}^m)\ \dd\tau \nonumber
\end{align}
As in equations (\ref{eq:node_loss}) and (\ref{eq:hn_loss_1}), $\mathbf{y}_t^m$ and $\hat{\mathbf{y}}_t^m$ denote the ground truth observation and the prediction for the $t^\text{th}$ timestep of task $m$ respectively, and $\bm{\uptheta}^m$ denotes the NODE parameters for task $m$ generated by the hypernetwork $\mathbf{f}_\mathbf{h}$ (with parameters $\mathbf{h}$) using the embedding vector $\mathbf{e}^m$ as input.
We use \myequation{eq:hn_loss_2} for training the hypernetwork in the second optimization step.
The structure of HN is shown in \myfigure{fig:arch_low}(e).
}
\smallskip

\noindent\bo{Chunked Hypernetworks (CHN)}: We use a chunked hypernetwork to generate the parameters of a NODE (\myfigure{fig:arch_low}(f)). For this, \myequation{eq:our_hn_loss_1} and \myequation{eq:hn_loss_2} are employed as the loss functions in the 2-step optimization process. 

\change{
To reproduce a particular task with HN or CHN at test time, we input the corresponding task embedding vector to the hypernetwork which generates the NODE for that task. Thereafter, we simply provide the NODE with the starting state of the trajectory (and the timesteps in the case of NODE$^\text{T}$) and it predicts the entire trajectory for the desired task.
}

\medskip

\new{
For the comparative baselines (SG, REP, FT, SI, MAS), we included representative methods from all the major groups of continual learning methods. The choice of hypernetworks for our proposed approach to continual LfD (HN, CHN) was motivated by the following: 
\begin{enumerate*}[label=(\roman*)]
	\item Hypernetworks have exhibited remarkable performance in multiple domains ~\citep{von2019continual, huang2021continual}.
	\item A robot capable of learning multiple tasks from demonstration needs to be told which task it should execute. Hypernetworks provide a natural way of task-conditioning using task embedding vectors, and thus are a very good fit for the LfD setup.
	\item Hypernetworks do not need to retrain on the data of past tasks.
	\item The chunked version of hypernetworks can be smaller than the NODEs they generate (model compression).
\end{enumerate*}
}


\section{Experiments}
\label{sec:experiments}

We first describe the datasets used in our experiments and the metrics we report. This is followed by the details of our experimental results.


\subsection{Datasets}
\label{sec:datasets}

\change{
We evaluate the performance of all models described in \mysection{sec:methods} on three different datasets with different numbers of tasks and degrees of difficulty:
\begin{enumerate*}[label=(\roman*)]
    \item \mbox{\textit{LASA}}~\citep{khansari2011learning}:~A dataset that is used frequently for comparisons and benchmarking in the area of LfD \citep{ravichandar2017learning, blocher2017learning, urain2020imitationflow};
    \item \mbox{\textit{HelloWorld}}:~A dataset of trajectories recorded while a robot was shown how to write letters containing loops;
    \item \mbox{\textit{RoboTasks}}:~A dataset of trajectories collected while the robot was shown how to perform tasks which need both the positions as well as orientations of the end-effector to be varied. 
\end{enumerate*}
Of these three datasets, \textit{HelloWorld} and \textit{RoboTasks} are introduced by us in this paper.
}

\subsubsection{LASA}
\label{sec:datasets_lasa}

LASA~\citep{khansari2011learning} is a widely-used benchmark
\change{\citep{ravichandar2017learning, blocher2017learning, urain2020imitationflow, saveriano2020energy}} for evaluating motion generation algorithms. It contains 30 patterns of handwritten motions, each with 7 similar demonstrations (see \myfigure{fig:lasa_traj} for examples).
We refer to each pattern $\mathcal{D}_{m}$ as a task. Of the 30 tasks, we use the first 26 tasks: $\bm{\mathcal{D}}_{\text{LASA}}=\{ \mathcal{D}_{0:25}\}$. We omit the last 4 tasks, each of which contains 2 or 3 dissimilar patterns merged together.
Each demonstration of a task is a sequence of 1000 2-D points.We arrange the 26 tasks alphabetically
in the order of their names and train sequentially models of SG, FT, REP, SI, MAS, CHN and HN on each task. During training only REP has access to the data of older tasks while the other methods are trained with the data of only the current task. 
\change{
We evaluate on the LASA dataset because of its wide adoption by the LfD community \citep{ravichandar2017learning, blocher2017learning, urain2020imitationflow, saveriano2020energy}, and because it contains a large number of diverse tasks which can be used to gauge the continual learning ability of our models.
}

\subsubsection{HelloWorld}
\label{sec:datasets_hw}

We further evaluate our approach on a dataset of demonstrations we collected using the Franka Emika Panda robot.
The $x$ and $y$ coordinates of the robot's end-effector were recorded while a human user guided it kinesthetically to write the $7$ lower-case letters \emph{h,e,l,o,w,r,d} one at a time on a horizontal surface. The \emph{HelloWorld} dataset $\bm{\mathcal{D}}_{\text{HW}}=\{ \mathcal{D}_{0:6}\}$ consists of $7$ tasks, each 
containing 8 slightly varying demonstrations of a letter. Each demonstration is a sequence of $1000$ 2-D points.
After training on all the tasks, the objective is to make the robot write the words ``$hello$ $world$''. 
\change{
Our motivation for using this dataset is to test our approach on trajectories containing loops and to show that it also works on kinesthetically recorded demonstrations using a real robot. 
}
This dataset is available at \url{https://github.com/sayantanauddy/clfd}.

\subsubsection{RoboTasks}
\label{sec:datasets_robtasks}

To evaluate our approach on robotic tasks that can be expected in the real world, we collect a dataset of 4 tasks. For each task,  a human user kinsethetically guides the robot's end-effector while varying both the position in 3D space and also the orientation in all three rotation axes. The tasks of this dataset are: 
\begin{enumerate*}[label=(\roman*)]
  \item \emph{box opening}: the lid of a box is lifted to an open position;
  \item \emph{bottle shelving}: a bottle in a vertical position is transferred to a horizontal position on a shelf;
  \item \emph{plate stacking}: a plate in a vertical position is transferred to a horizontal position on an elevated platform while orienting the arm so as to avoid the blocks used for holding the plate in its initial vertical position;
  \item \emph{pouring}: a cup full of coffee beans is taken from an elevated platform and the contents of the cup are emptied into a container.
\end{enumerate*}
Thus, the \emph{RoboTasks} dataset $\bm{\mathcal{D}}_{\text{Robot}}=\{ \mathcal{D}_{0:3}\}$ consists of 4 tasks, where each task contains 9 demonstrations, each of 1000 steps. In each step we record the position $\mathbf{p} \in \mathbb{R}^3$ and the orientation as a unit quaternion $\mathbf{q} \in \mathbb{S}^3$. Compared to $\bm{\mathcal{D}}_{\text{LASA}}$ and $\bm{\mathcal{D}}_{\text{HW}}$, $\bm{\mathcal{D}}_{\text{Robot}}$ contains a lot more variability in the demonstrations and presents a more difficult learning challenge.

\change{
We use this dataset because we want to evaluate our approach in a \change{real-world} setup involving changing positions and orientations, and also because we did not find any existing datasets that are suitable for our purpose. Datasets such as RoboNet \citep{robotnet} and Meta-World \citep{metaworld} are designed for reinforcement learning and do not contain the demonstrations in terms of task-space trajectories.}
See \myfigure{fig:robottasks_prediction} for a visual example of the tasks in $\bm{\mathcal{D}}_{\text{Robot}}$. This dataset is also available at \url{https://github.com/sayantanauddy/clfd}.


\subsection{Metrics}
\label{sec:metrics}

\subsubsection{Trajectory Metrics}
\label{sec:metrics_traj}

To evaluate the end-effector position trajectories, we report the following widely used metrics: 
\emph{Dynamic Time Warping error} (DTW) \citep{urain2020imitationflow, jekel2019similarity}, \emph{Swept Area error} \citep{khansari2014learning}, and \emph{Frechet distance} \citep{urain2020imitationflow}, which measure how close the predicted trajectories are to the ground-truth demonstrations. For orientation, we report the commonly used quaternion error \citep{saveriano2019merging, ude2014orientation} defined as 
\begin{equation}
	e_q(\mathbf{q}_1,\mathbf{q}_2)  = 2\logmap(\mathbf{q}_1 * \bar{\mathbf{q}}_2) \in \mathbb{R}^3
	\label{eq:quat_err1}
\end{equation}
where $\mathbf{q}_1$ and $\mathbf{q}_2$ are quaternions. Given a ground truth trajectory $\mathbf{q}_{0:T-1}$ and a predicted trajectory $\hat{\mathbf{q}}_{0:T-1}$ (both containing $T$ quaternions), we compute the error between the two trajectories using
\begin{equation}
	E_q = \cfrac{1}{3T}\sum_{t=0}^{T-1}||e_q(\mathbf{q}_t,\hat{\mathbf{q}}_t) ||_{1}
	\label{eq:quat_err2}
\end{equation}

\subsubsection{Continual Learning Metrics}
\label{sec:metrics_cl}

We report \emph{Accuracy} (ACC), \emph{Remembering} (REM), \emph{Model Size Efficiency} (MS) and \emph{Sample Storage Size Efficiency} (SSS)~\citep{diaz2018don}.
ACC is a measure of the average accuracy for the current and past tasks. REM measures how well past tasks are remembered. MS measures how much the size of a model grows compared to its size after learning the first task. SSS measures how the storage requirements of a model grows due to the storage of training data from older tasks. 

Additionally, we introduce two new easy-to-compute continual learning metrics: 
\emph{Time Efficiency} (TE) and \emph{Final Model Size} (FS). 
TE measures the increase in training duration with the number of tasks, relative to the training time for the first task. TE only needs the training times to be logged, and it reflects the extra effort needed in the training loop (e.g. due to extra regularization steps) with an increase in the number of tasks. 
\new{
TE is similar to the \emph{Computational Efficiency} metric (CE) proposed in \cite{diaz2018don}, but while CE is based on the number of addition and multiplication operations, TE is based on the observed wall clock time. When a neural network is trained on a GPU (as is most often the case), many neural network operations are performed in parallel. Hence, simply counting the number of operations does not provide an accurate estimate of the increase in training effort with increasing number of tasks. A time-based measure such as TE is more suited to this task. For a meaningful interpretation of this metric, all experiments need to be run on near-identical hardware. 
}

For $M$ tasks, TE is defined as
\begin{equation}
	\textrm{TE}=\min\left\{1, \frac{\mathcal{T}_0}{M} \sum_{i=0}^{M-1} \frac{1}{\mathcal{T}_i} \right\},
\end{equation}
where $\mathcal{T}_i$ is the time required to learn task $i$. 

FS is a measure of the \emph{absolute} parameter size, which contrasts with MS which only measures the parameter growth relative to the size after learning the first task. A model which has a large number of parameters for the first task and adds a relatively small number of parameters for subsequent tasks will achieve a high score for MS, but will fare worse in terms of FS if models of other compared methods have a smaller absolute size. FS is defined as
\begin{equation}
	\textrm{FS} = 1-\mathrm{Mem}_\text{norm}(\bm{\uptheta}^{M-1})
\end{equation}
$\mathrm{Mem}_\text{norm}(\bm{\uptheta}^{M-1})$ is the parameter size after learning $M$ tasks, normalized by the size of the largest compared model among SG, FT, REP, SI ,MAS, HN and CHN. 
With these 6 metrics, we compute the overall continual learning metrics \cite{diaz2018don}: 
\begin{align}
&\textrm{CL}_{\text{score}} = \small\sum^{n(\mathcal{C})}c_i, \\
& \textrm{CL}_{\text{stability}} = 1- \sum^{n(\mathcal{C})}\text{STDEV}(c_i),
\end{align}

where $\mathcal{C}=\{\textrm{ACC, REM, MS, TE, FS, SSS}\}$. 
All the continual learning metrics lie in the range 0 (worst) to 1 (best).


\subsection{Results}
\label{sec:results}

\new{
We present next the results of our experiments on three different datasets: \emph{LASA}, \emph{HelloWorld}, and \emph{RoboTasks}. 
Details of our hardware setup (\mysection{sec:appendix_hardware}), experiment hyperparameters (\mysection{sec:appendix_hyperparameters}), results of the experimental comparison of NODEs and iFlow (\mysection{sec:iflow}), and additional results (\mysection{sec:appendix_additional_res}) can be found in the appendix.
}


\subsubsection{LASA dataset}
\label{sec:res_lasa}

\begin{figure}[t!]
	\includegraphics[width=\textwidth]{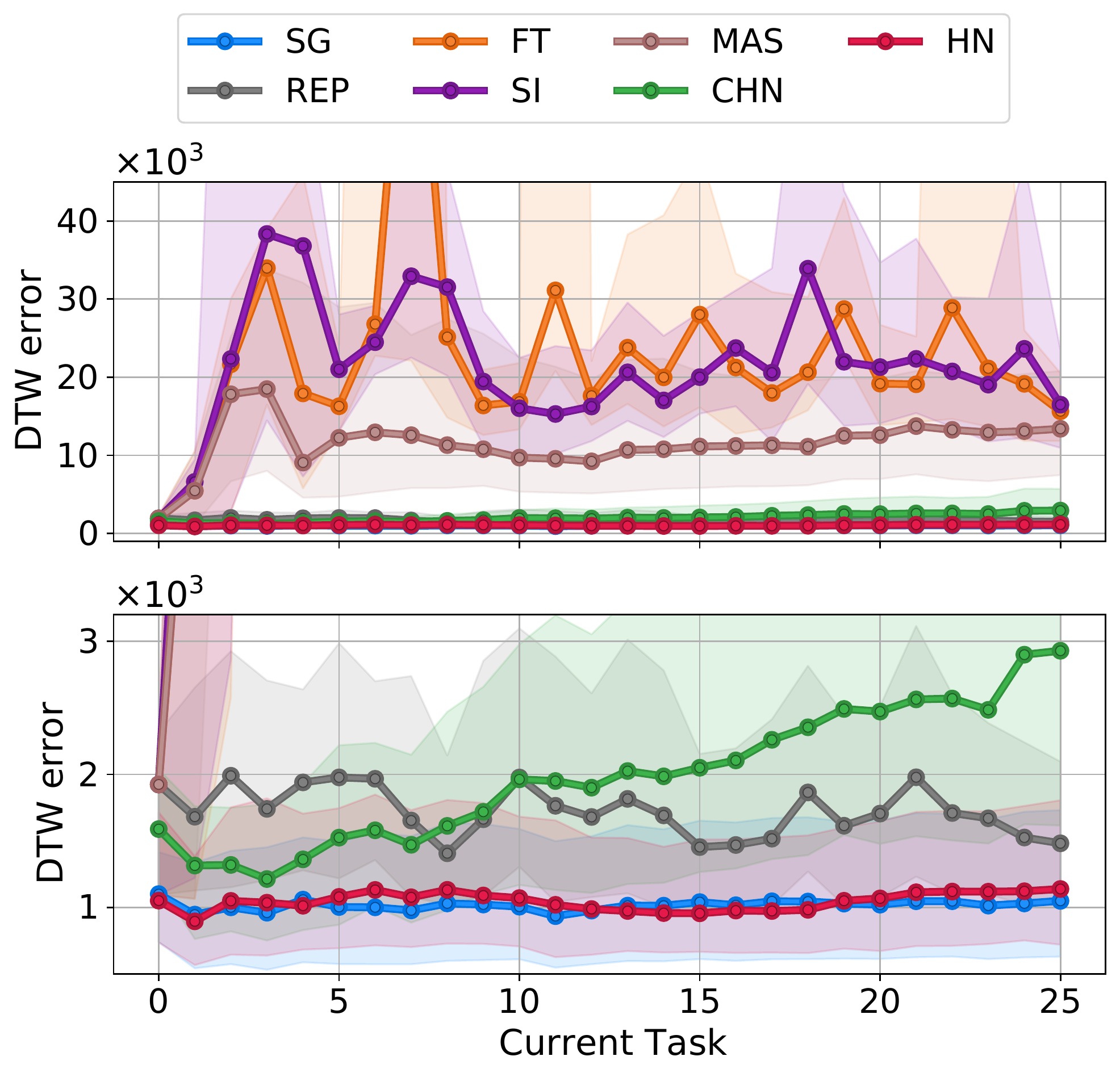}
	\caption{DTW errors of trajectories predicted for the LASA dataset (lower is better).
	The x-axis shows the \emph{current} task. After learning a task (using NODE$^\text{T}$), all current and previous tasks are evaluated. HN performs as well as SG but with negligible growth in parameter size. Lines show medians and shaded regions denote the lower and upper quartiles of the errors over 5 independent seeds. (top) The DTW errors for all methods. (bottom) A zoomed-in view of the methods that perform well. }
	\label{fig:lasa_cumu}
\end{figure}

We train each model (SG, FT, REP, SI, MAS, CHN and HN) on the $26$ tasks of $\bm{\mathcal{D}}_{\text{LASA}}$ sequentially. \myfigure{fig:lasa_cumu} shows the median DTW errors of the predictions for tasks $\mathcal{D}_0$--$\mathcal{D}_m$ after training on task $\mathcal{D}_m$ (using NODE$^{\text{T}}$) for $m=0,1,\ldots25$. For example, the value for task 10 denotes the median of the evaluation errors for all the predicted trajectories from tasks 0 to 10 after training on task 10. We provide the same information using the other trajectory error metrics (Frechet distance and Swept Area in \myfigure{fig:lasa_cumu_all_metrics} in the appendix).

\myfigure{fig:lasa_cumu} (top) shows a drastic increase in the error for FT as the number of learned tasks increases, since FT optimizes its parameters \emph{only} for the current task. After the first task, the errors for SI and MAS also increase steeply.
\myfigure{fig:lasa_cumu} (bottom) shows a zoomed-in version showing the methods which perform well in more detail (SG, REP, CHN and HN).
The performance of REP for task 0 is the same as FT, but the performance does not deteriorate as more tasks are learned, since REP has access to the data from prior tasks.
Till task~7, CHN performs better than REP, after which it starts exhibiting catastrophic forgetting as can be seen from the upward trend in its error plot.
Overall, SG and HN exhibit the best performance. They do not suffer from catastrophic forgetting, and their error plots overlap with each other (red and blue lines in \myfigure{fig:lasa_cumu} (bottom)). 

\begin{figure}[t!]
	\includegraphics[width=\textwidth]{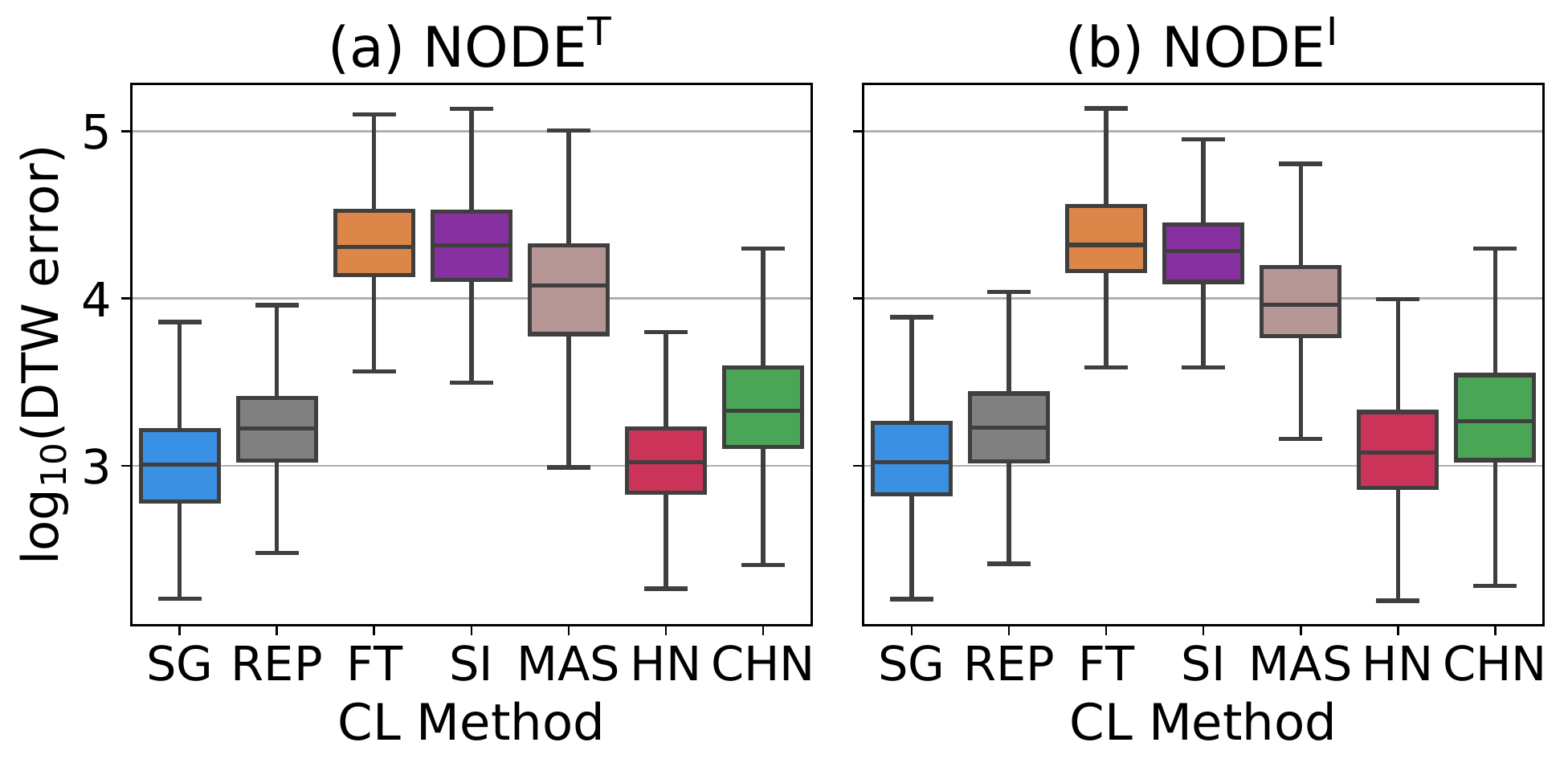}
	\caption{DTW errors (lower is better) of all trajectory predictions during the learning of all tasks of the LASA dataset. Results are obtained using 5 independent seeds. The hypernetwork approach (HN, CHN) performs substantially better than regularization-based continual learning methods (SI, MAS) and on par with the upper baseline SG and REP .
	}
	\label{fig:lasa_box}
\end{figure}

An overall picture of the continual learning performance of the different methods may be obtained from \myfigure{fig:lasa_box}, which shows the overall errors of the predicted trajectories during the course of learning all the tasks. 
It can be seen that SG, REP, CHN and HN perform much better than FT, SI, and MAS, and that HN and SG are the best performers. Note that in \myfigure{fig:lasa_box} the trajectory metrics are plotted in the $\log_{10}$ scale to accommodate the high errors for FT, SI, and MAS in the same plot as SG, REP, HN and CHN.
\begin{figure}[b!]
	\includegraphics[width=\textwidth]{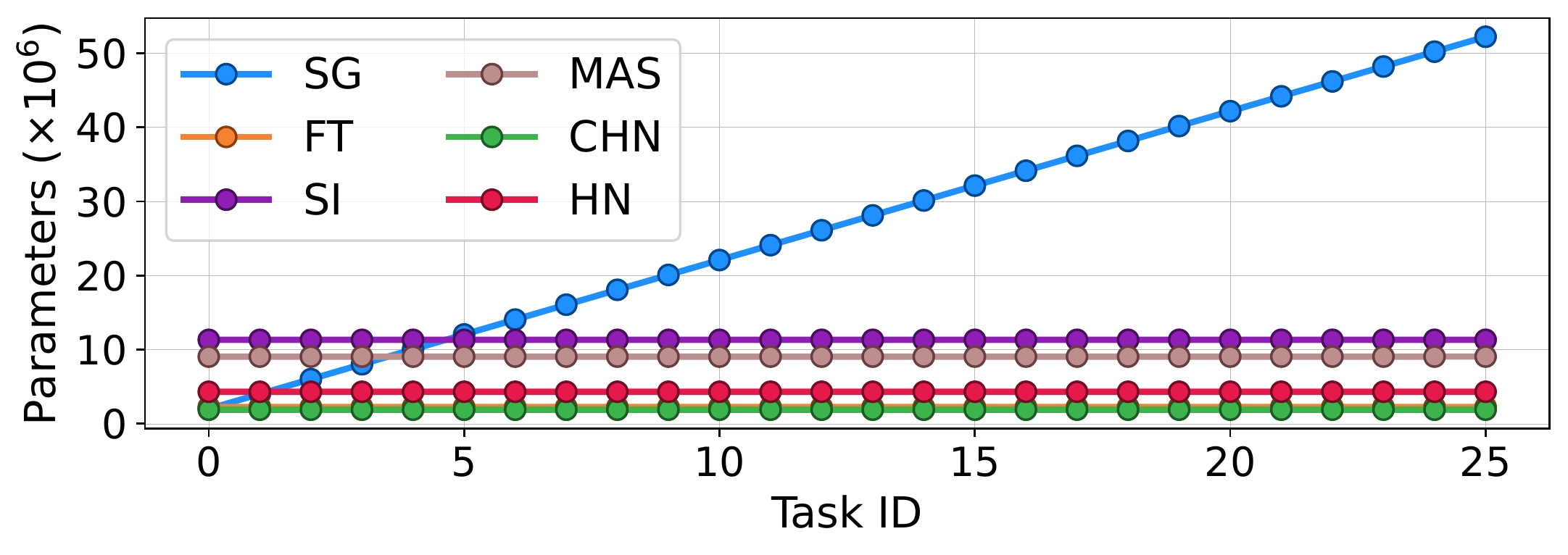}
	\caption{Growth of parameter size with new tasks for the LASA dataset (using NODE$^\text{T}$). SG has a high rate of growth since it uses a separate network for each task. All other models grow by only 256 parameters for each new task. Plots for CHN and FT overlap with each other. The plot for REP is not shown since its size is identical to FT.}
	\label{fig:lasa_model_size}
\end{figure}

\begin{figure}[b!]
	\includegraphics[width=\textwidth]{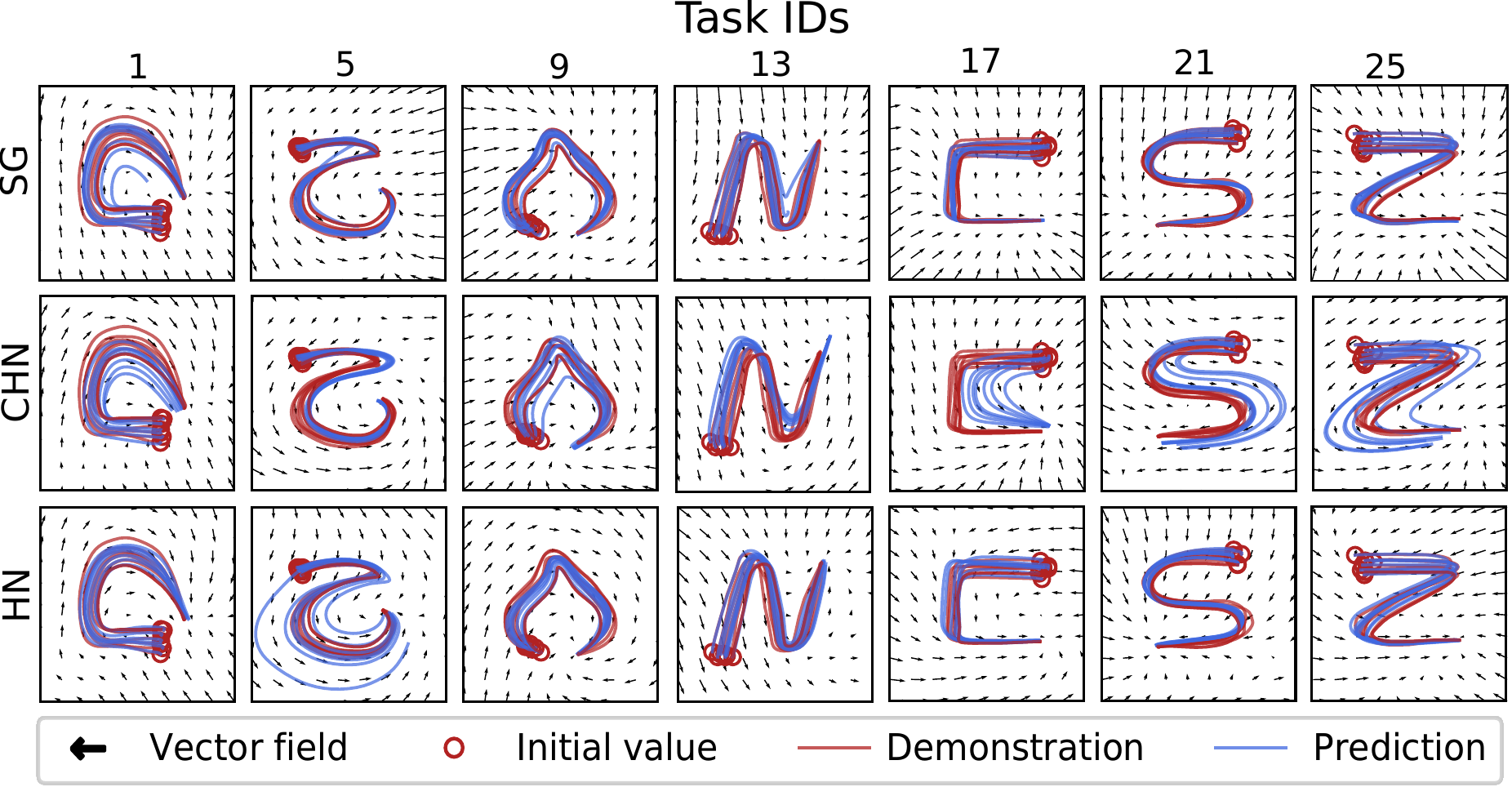}
	\caption{Example of trajectories predicted by SG, CHN and HN using NODE$^{\text{T}}$ for a selection of LASA tasks after learning the last task. HN can remember all past tasks, while CHN exhibits some forgetting.}
	\label{fig:lasa_traj}
\end{figure}

\change{Note that a regular HN model produces all the parameters of the target NODE directly from its final layer. This implies a much larger number of HN parameters compared to the produced target network parameters. To keep the overall size of the HN comparable to the other models, the target NODE for HN has one-tenth the number of units in each layer as the NODEs used by the other models (SG, FT, REP, SI, MAS, CHN). 
We refer the reader to \mytable{tab:hparam_lasa} in the appendix for details.}

\begin{figure*}[b!]
	\centering
	\includegraphics[width=\textwidth]{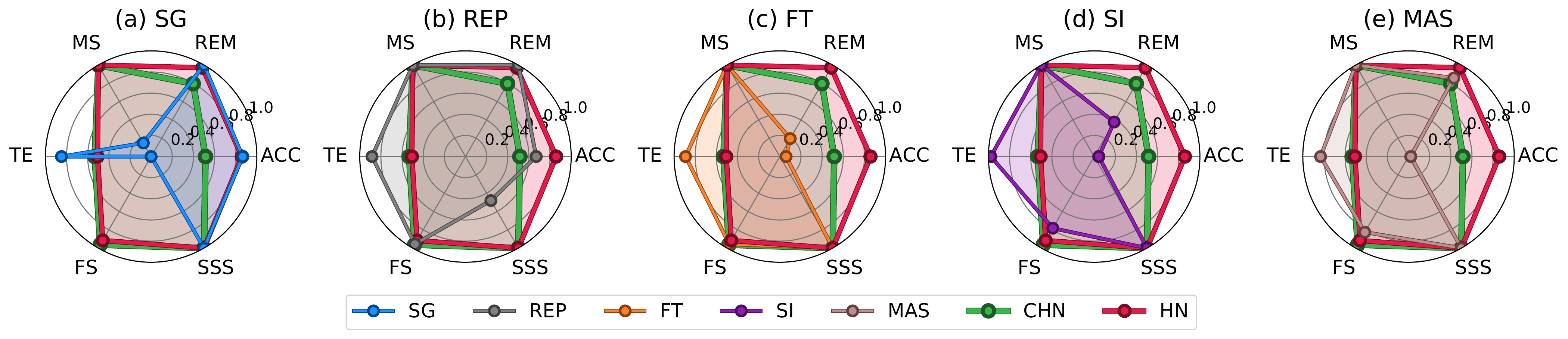}
	\caption{
	\new{Continual learning metrics (0: worst, 1: best) for the different methods (with NODE$^\mathrm{T}$) on the LASA dataset. In (a)-(e), each of the baseline methods is compared against HN and CHN to  illustrate the different tradeoffs made by the methods.
	HN shows good performance for almost all metrics (apart from TE, in which it performs moderately). CHN is similar to HN in terms of TE but scores lower in REM and ACC. REP also performs well but has a low score for SSS, and the other methods fail in multiple metrics.
	}
	}
	\label{fig:spider_lasa}
\end{figure*}

Although HN's performance after learning $26$ tasks is very similar to that of SG, its parameter size is $4.3\times 10^6$ compared to SG's combined size of $52.2\times10^6$ parameters, as shown in \myfigure{fig:lasa_model_size}. CHN's  final parameter size is $1.9 \times 10^6$. Also, the parameter count for SG grows by $2.1 \times 10^6$ per task, whereas  CHN and HN grow at a much smaller rate of $256$ parameters per task, same as SI, MAS, FT and REP (\myfigure{fig:lasa_model_size} does not include REP as it has the same model size as FT). 
Thus, CHN and especially HN perform similar to the upper baseline SG, while their parameter size is close to the lower baseline FT.
\myfigure{fig:lasa_traj} shows  examples of trajectories predicted by SG, CHN, and HN for a selection of past tasks after learning the last task. 

To compute the continual learning metrics~\citep{diaz2018don}, each predicted trajectory needs to be marked as accurate or inaccurate based on its difference to the ground truth. 
Since there is no preexisting procedure for this, we adopt the following 
approach: We set a threshold on the DTW error, such that predictions with an error less than the threshold are considered accurate. 
As each task has multiple ground truth demonstrations, we first compute the DTW error between all pairs of demonstrations for each task.
\new{
We then find the maximum value from this list and multiply it by 3.
The multiplicative factor 3 was chosen empirically to allow some room for error such that a predicted trajectory with the same general shape as its demonstration (discerned visually) is considered accurate. Doing so, we arrive at a DTW threshold of $2191$ for $\bm{\mathcal{D}}_{\text{LASA}}$, and use it to evaluate the continual learning metrics shown in \mytable{tab:lasa_cl}. 
}

\begin{table}[t!]
	\caption{Continual learning metrics for the LASA dataset (median over 5 seeds). Values range from 0 (worst) to 1 (best).}
	\subfloat[NODE$^{\text{T}}$]{
	\centering
	\resizebox{\textwidth}{!}{
	\begin{tabular}{lllllllll}
\toprule
\textbf{MET} &   \textbf{ACC} &   \textbf{REM} &    \textbf{MS} &    \textbf{TE} &    \textbf{FS} &   \textbf{SSS} & \textbf{CL$_{sco}$} & \textbf{CL$_{stab}$} \\
\midrule
          SG &  \textbf{0.87} &  \textbf{1.00} &           0.15 &           0.85 &           0.00 &  \textbf{1.00} &                0.64 &                 0.55 \\
          FT &           0.06 &           0.20 &  \textbf{1.00} &           0.89 &  \textbf{0.96} &  \textbf{1.00} &                0.68 &                 0.57 \\
         REP &           0.67 &  \textbf{1.00} &  \textbf{1.00} &           0.88 &  \textbf{0.96} &           0.48 &                0.83 &                 0.79 \\
          SI &           0.04 &           0.38 &  \textbf{1.00} &  \textbf{0.98} &           0.78 &  \textbf{1.00} &                0.70 &                 0.60 \\
         MAS &           0.02 &           0.86 &  \textbf{1.00} &           0.83 &           0.83 &  \textbf{1.00} &                0.76 &                 0.63 \\
          HN &           0.86 &           0.97 &  \textbf{1.00} &           0.51 &           0.92 &  \textbf{1.00} &       \textbf{0.88} &        \textbf{0.81} \\
         CHN &           0.51 &           0.80 &  \textbf{1.00} &           0.53 &  \textbf{0.96} &  \textbf{1.00} &                0.80 &                 0.77 \\
\bottomrule
\end{tabular}

	}
	\label{tab:lasa_cl_explicit_time}
	}

	\subfloat[NODE$^{\text{I}}$]{
	\centering
	\resizebox{\textwidth}{!}{
	\begin{tabular}{lllllllll}
\toprule
\textbf{MET} &   \textbf{ACC} &   \textbf{REM} &    \textbf{MS} &    \textbf{TE} &    \textbf{FS} &   \textbf{SSS} & \textbf{CL$_{sco}$} & \textbf{CL$_{stab}$} \\
\midrule
          SG &  \textbf{0.81} &  \textbf{1.00} &           0.15 &           0.85 &           0.00 &  \textbf{1.00} &                0.64 &                 0.56 \\
          FT &           0.06 &           0.22 &  \textbf{1.00} &           0.88 &  \textbf{0.96} &  \textbf{1.00} &                0.69 &                 0.57 \\
         REP &           0.65 &  \textbf{1.00} &  \textbf{1.00} &           0.83 &  \textbf{0.96} &           0.48 &                0.82 &                 0.79 \\
          SI &           0.04 &           0.38 &  \textbf{1.00} &  \textbf{0.91} &           0.78 &  \textbf{1.00} &                0.69 &                 0.61 \\
         MAS &           0.03 &           0.84 &  \textbf{1.00} &           0.87 &           0.83 &  \textbf{1.00} &                0.76 &                 0.63 \\
          HN &           0.76 &           0.97 &  \textbf{1.00} &           0.55 &           0.92 &  \textbf{1.00} &       \textbf{0.87} &        \textbf{0.82} \\
         CHN &           0.57 &           0.86 &  \textbf{1.00} &           0.52 &  \textbf{0.96} &  \textbf{1.00} &                0.82 &                 0.78 \\
\bottomrule
\end{tabular}

	}
	\label{tab:lasa_cl_no_time}
	}	
\label{tab:lasa_cl}
\end{table}

For both NODE variants, HN outperforms all the compared models in terms of CL$_{\text{score}}$. For NODE$^{\text{T}}$, HN performs close to the upper baseline SG in terms of both ACC and REM. 
The additional regularization needed for training hypernetworks leads to a comparatively lower score for the time efficiency metric TE for CHN and HN (actual wall-clock training times can be found in \mytable{tab:wall_clock}(a) in the appendix). A very high parameter growth rate for SG results in poor scores for MS and FS. The replay method REP achieves a low score for SSS since it is the only method that needs to store the training data from previous tasks. The direct time input in NODE$^\text{T}$ also leads to better ACC scores for SG, REP and HN, compared to NODE$^\text{I}$. \new{
\myfigure{fig:spider_lasa} shows how the different methods measure up against each other in the different aspects of continual learning performance.
}

\begin{table}[b!]
\caption{Robustness to changes in regularization hyperparameters for the LASA dataset (5 configurations for each method)}
\centering
\resizebox{0.8\textwidth}{!}{
	\begin{tabular}{l@{\hskip 15pt}rr@{\hskip 15pt}rr}
	       &   \multicolumn{2}{c}{CL$_\text{score}$} & \multicolumn{2}{c}{CL$_\text{stability}$}\\
	\toprule
    METHOD &       Median &           IQR &           Median &               IQR \\
   \midrule
        HN &       0.8578 &        0.0011 &           0.8324 &            0.0050 \\
       CHN &       0.7939 &        0.0098 &           0.8126 &            0.0022 \\
       MAS &       0.7104 &        0.0019 &           0.6562 &            0.0062 \\
        SI &       0.6047 &        0.0065 &           0.6403 &            0.0011 \\
   \bottomrule
	\end{tabular}
	}
\label{tab:lasa_hyperparam_study}
\end{table}

To test the sensitivity of the regularization-based continual learning methods (SI, MAS, HN and CHN) to changes in the regularization hyperparameters, we create sets of 5 hyperparameters each for SI, MAS, HN and CHN by drawing independently and uniformly from the following ranges: (SI) $c \in [0.1,0.5], \xi \in [0.1, 0.5]$, (MAS) $c \in [0.1,0.5]$, (CHN) $\beta \in [10^{-3}, 10^{-2}]$, (HN) $\beta \in [10^{-3}, 10^{-2}]$ resulting in 20 different configurations. We then repeat the LASA experiment with NODE$^{\text{T}}$ for all these configurations. In terms of CL$_\text{score}$ we observe that all configurations of HN outperform all configurations of CHN, which in turn are better than all configurations of MAS, followed by SI. This trend is reflected in the medians and inter-quartile ranges (IQR) of the overall continual learning metrics CL$_\text{score}$ and CL$_\text{stability}$ for each method (over its 5 configurations) shown in \mytable{tab:lasa_hyperparam_study}. It can be seen that HN and CHN perform better than the other methods and the variability in terms of IQR is very small, thereby showing that they are robust to changes in the regularization hyperaparameter $\beta$.  For calculating CL$_\text{score}$ and CL$_\text{stability}$ in this experiment, we do not consider the SSS metric since none of the regularization-based methods need to cache training data from prior tasks.

\begin{figure}[b!]
	\setcounter{figure}{8}
	\includegraphics[width=\textwidth]{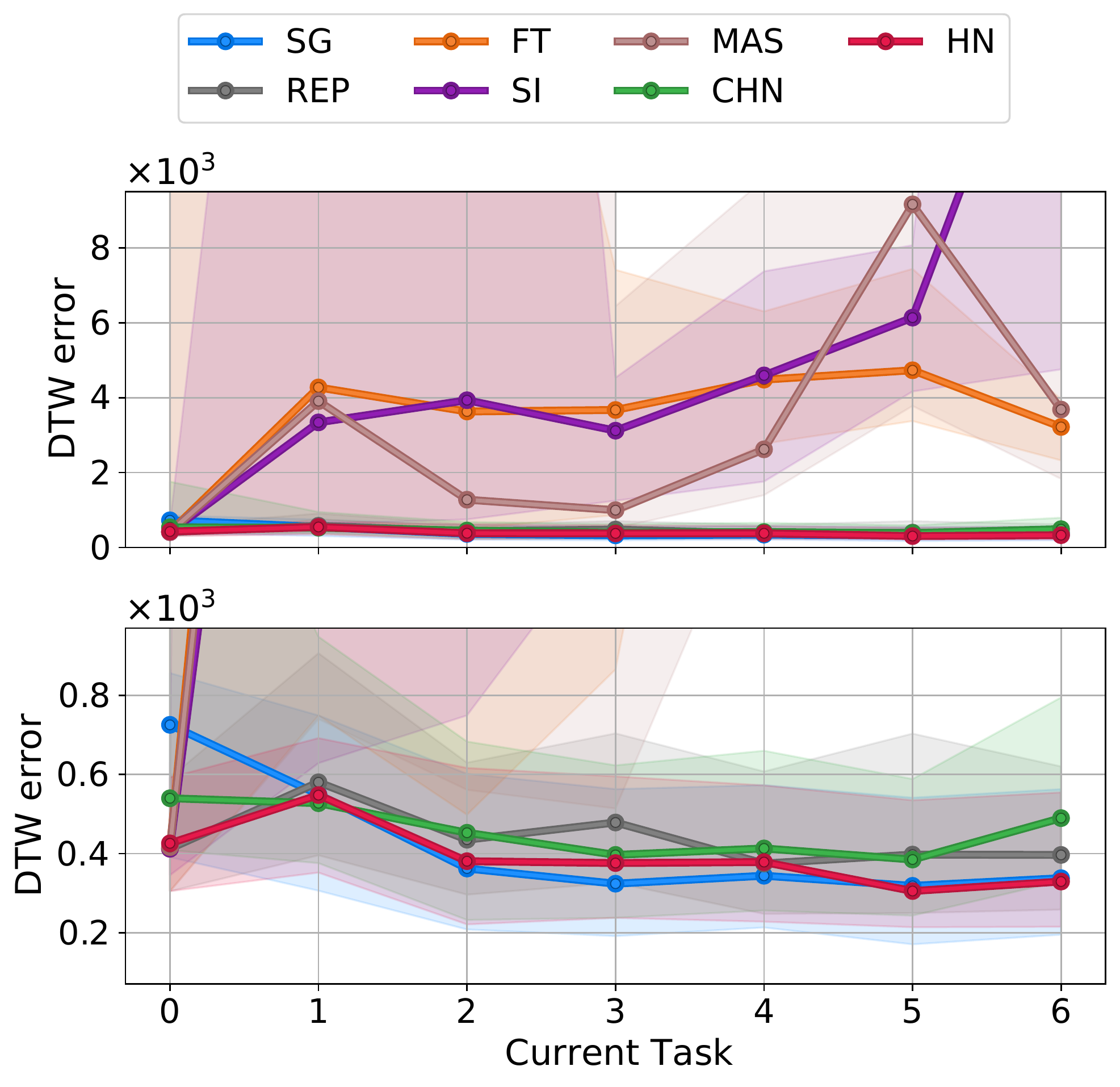}
	\caption{DTW errors of trajectories predicted for the HelloWorld dataset (lower is better). The x-axis shows the \emph{current} task. After learning a task (using NODE$^\text{T}$), all current and previous tasks are evaluated. Lines show medians and shaded regions denote the lower and upper quartiles of the errors over 5 independent seeds. (top) The DTW errors for all methods. (bottom) A zoomed-in view of the methods that perform well. The performance of SG, REP, HN and CHN are nearly identical and much better than FT, SI and MAS.}
	\label{fig:hw_cumu}
\end{figure}

\begin{figure*}[b!]
	\setcounter{figure}{11}
	\centering
	\includegraphics[width=\textwidth]{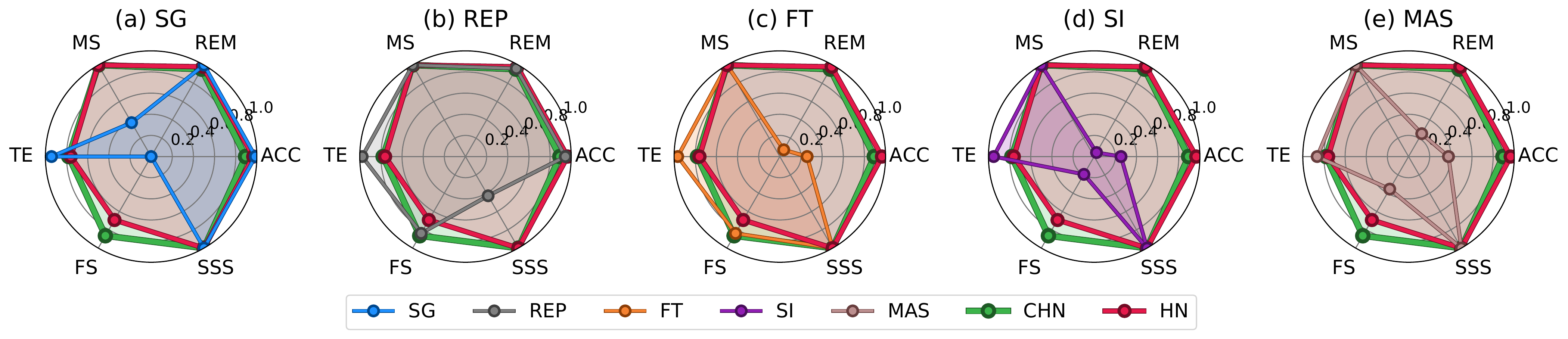}
	\caption{\new{Continual learning metrics (0: worst, 1: best) for the different methods (with NODE$^\mathrm{T}$) on the HelloWorld dataset. In (a)-(e), each of the baseline methods is compared against HN and CHN to  illustrate the different tradeoffs made by the methods.
	The hypernetwork methods (HN,CHN) show good performance for all metrics whereas the other methods perform well only in some metrics and poorly for others.
	}
	}
	\label{fig:spider_hw}
\end{figure*}


\begin{figure}[t!]
	\setcounter{figure}{9}
	\includegraphics[width=0.95\textwidth]{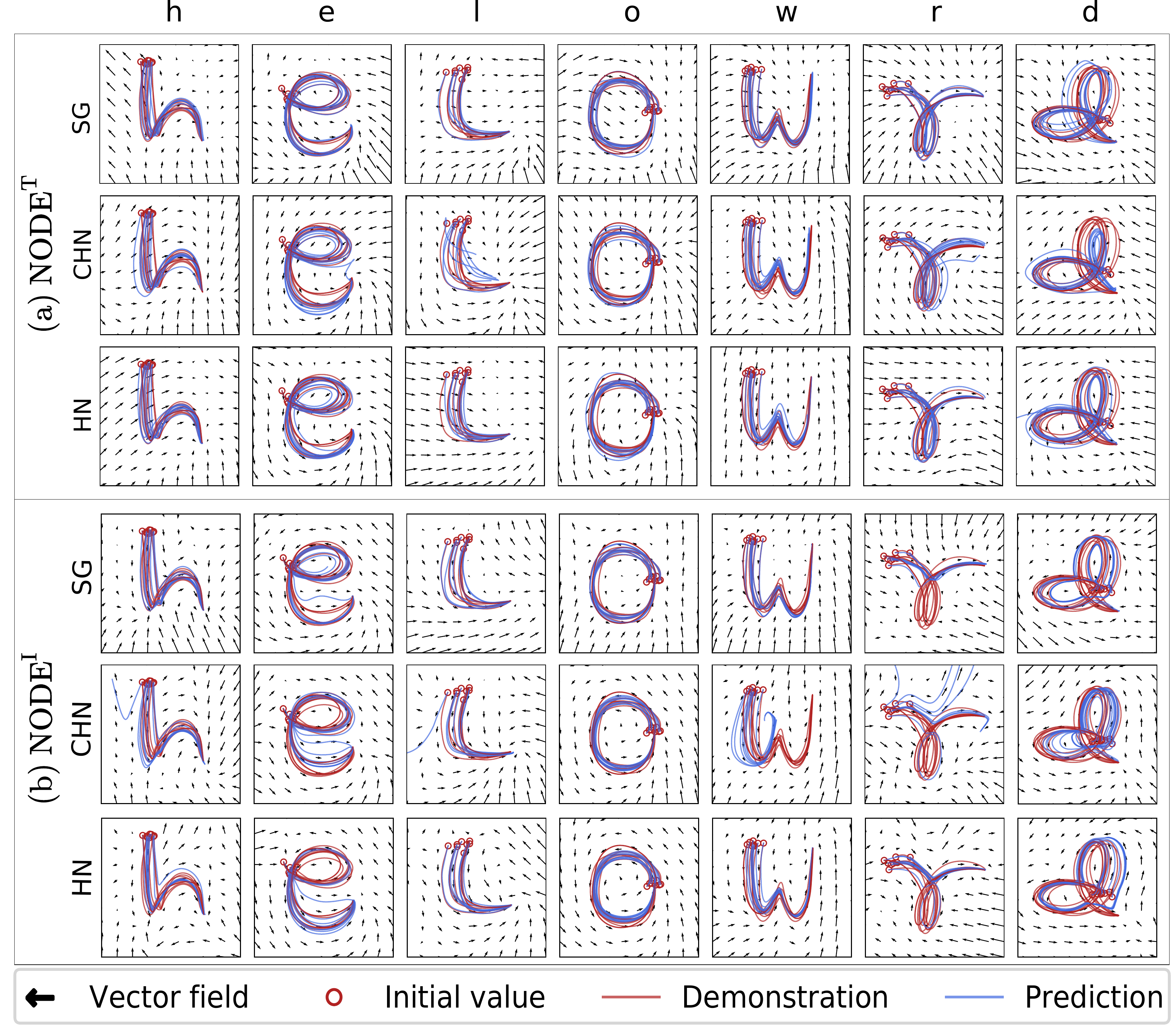}
	\caption{Example of trajectories predicted by SG, CHN and HN using NODE$^{\text{T}}$ for all past HelloWorld tasks after learning the last task. With NODE$^{\mathrm{T}}$, even trajectories with loops can be learned. HN and CHN (with NODE$^{\mathrm{T}}$) can remember all past tasks.}
	\label{fig:hw_traj}
\end{figure}
\begin{figure}[t!]
	\includegraphics[width=\textwidth]{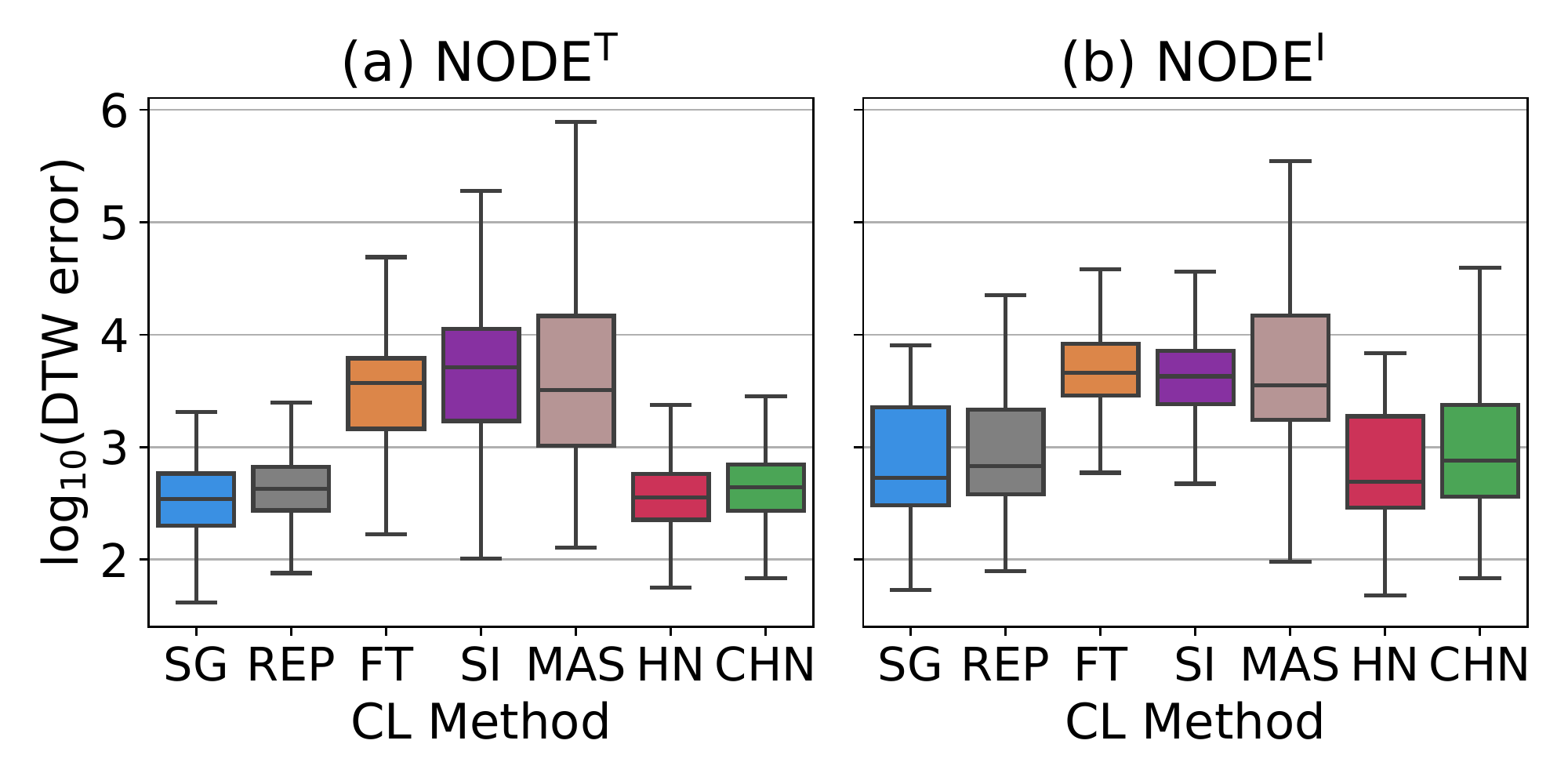}
	\caption{DTW errors (lower is better) of all trajectory predictions during the learning of all tasks of the HelloWorld dataset. Results are obtained using 5 independent seeds. HN, CHN perform as well as the upper baseline SG.}
	\label{fig:hw_box}
\end{figure}

\subsubsection{HelloWorld}
\label{sec:res_hw}

For $\bm{\mathcal{D}}_{\text{HW}}$, which comprises $7$ tasks, we perform the same experiments as $\bm{\mathcal{D}}_{\text{LASA}}$. \myfigure{fig:hw_cumu} shows the errors in the predicted trajectories for all past and current tasks, as new tasks are learned. The median errors for CHN, HN and REP stay nearly unchanged and are similar to the upper baseline SG, although CHN and HN do not need to store the training data of past tasks like REP. As before, FT, SI, and MAS exhibit severe catastrophic forgetting.
Due to fewer tasks in $\bm{\mathcal{D}}_{\text{HW}}$, CHN's performance does not deteriorate even after learning all tasks. \myfigure{fig:hw_cumu_all_metrics} in the appendix shows the other trajectory error metrics (Frechet distance and Swept Area), which exhibit the same trend as DTW.

\myfigure{fig:hw_traj} shows examples of trajectories predicted by SG, CHN and HN for past tasks after being trained sequentially on all $\bm{\mathcal{D}}_{\text{HW}}$ tasks. All models exhibit superior performance when using the additional time input in NODE$^\text{T}$ (\myfigure{fig:hw_traj}(a)), without which even SG is unable to learn trajectories with loops. This can be seen from the errors for the letters \emph{e},~\emph{r}~and~\emph{d} in \myfigure{fig:hw_traj}(b). 
This is also evident in \myfigure{fig:hw_box} which shows the errors for all predictions during the course of learning all the tasks. Apart from FT, all methods have higher median errors when using NODE$^{\text{I}}$ (\myfigure{fig:hw_box}(b)) compared to NODE$^{\text{T}}$ (\myfigure{fig:hw_box}(a)). 

\begin{table}[t!]
	\caption{Continual learning metrics for the HelloWorld dataset (median over 5 seeds). Values range from 0 (worst) to 1 (best).}
\subfloat[NODE$^{\text{T}}$]{
	\centering
	\resizebox{\textwidth}{!}{
	\begin{tabular}{lllllllll}
\toprule
\textbf{MET} &   \textbf{ACC} &   \textbf{REM} &    \textbf{MS} &    \textbf{TE} &    \textbf{FS} &   \textbf{SSS} & \textbf{CL$_{sco}$} & \textbf{CL$_{stab}$} \\
\midrule
          SG &  \textbf{0.99} &  \textbf{1.00} &           0.37 &           0.94 &           0.00 &  \textbf{1.00} &                0.72 &                 0.57 \\
          FT &           0.26 &           0.07 &  \textbf{1.00} &           0.97 &           0.84 &  \textbf{1.00} &                0.69 &                 0.59 \\
         REP &           0.94 &           0.96 &  \textbf{1.00} &  \textbf{0.98} &           0.84 &           0.43 &                0.86 &                 0.78 \\
          SI &           0.25 &           0.04 &  \textbf{1.00} &           0.95 &           0.19 &  \textbf{1.00} &                0.57 &                 0.55 \\
         MAS &           0.38 &           0.25 &  \textbf{1.00} &           0.86 &           0.36 &  \textbf{1.00} &                0.64 &                 0.65 \\
          HN &           0.97 &           0.98 &  \textbf{1.00} &           0.76 &           0.69 &  \textbf{1.00} &                0.90 &                 0.86 \\
         CHN &           0.89 &           0.95 &  \textbf{1.00} &           0.78 &  \textbf{0.87} &  \textbf{1.00} &       \textbf{0.92} &        \textbf{0.91} \\
\bottomrule
\end{tabular}

	}
	\label{tab:hw_cl_explicit_time}
	}

	\subfloat[NODE$^{\text{I}}$]{
	\centering
	\resizebox{\textwidth}{!}{
	\begin{tabular}{lllllllll}
\toprule
\textbf{MET} &   \textbf{ACC} &   \textbf{REM} &    \textbf{MS} &    \textbf{TE} &    \textbf{FS} &   \textbf{SSS} & \textbf{CL$_{sco}$} & \textbf{CL$_{stab}$} \\
\midrule
          SG &           0.71 &  \textbf{1.00} &           0.37 &           0.93 &           0.00 &  \textbf{1.00} &                0.67 &                 0.59 \\
          FT &           0.17 &           0.24 &  \textbf{1.00} &           0.93 &           0.84 &  \textbf{1.00} &                0.70 &                 0.62 \\
         REP &           0.71 &  \textbf{1.00} &  \textbf{1.00} &  \textbf{0.94} &           0.84 &           0.43 &                0.82 &                 0.78 \\
          SI &           0.19 &           0.29 &  \textbf{1.00} &  \textbf{0.94} &           0.19 &  \textbf{1.00} &                0.60 &                 0.59 \\
         MAS &           0.25 &           0.47 &  \textbf{1.00} &           0.86 &           0.36 &  \textbf{1.00} &                0.66 &                 0.66 \\
          HN &  \textbf{0.74} &           0.99 &  \textbf{1.00} &           0.75 &           0.69 &  \textbf{1.00} &                0.86 &                 0.85 \\
         CHN &           0.68 &           0.96 &  \textbf{1.00} &           0.76 &  \textbf{0.87} &  \textbf{1.00} &       \textbf{0.88} &        \textbf{0.87} \\
\bottomrule
\end{tabular}

	}
	\label{tab:hw_cl_no_time}
	}
\label{tab:hw_cl}
\end{table}

\begin{figure}[t]
	\setcounter{figure}{12}
	\centering
	\includegraphics[width=0.92\textwidth]{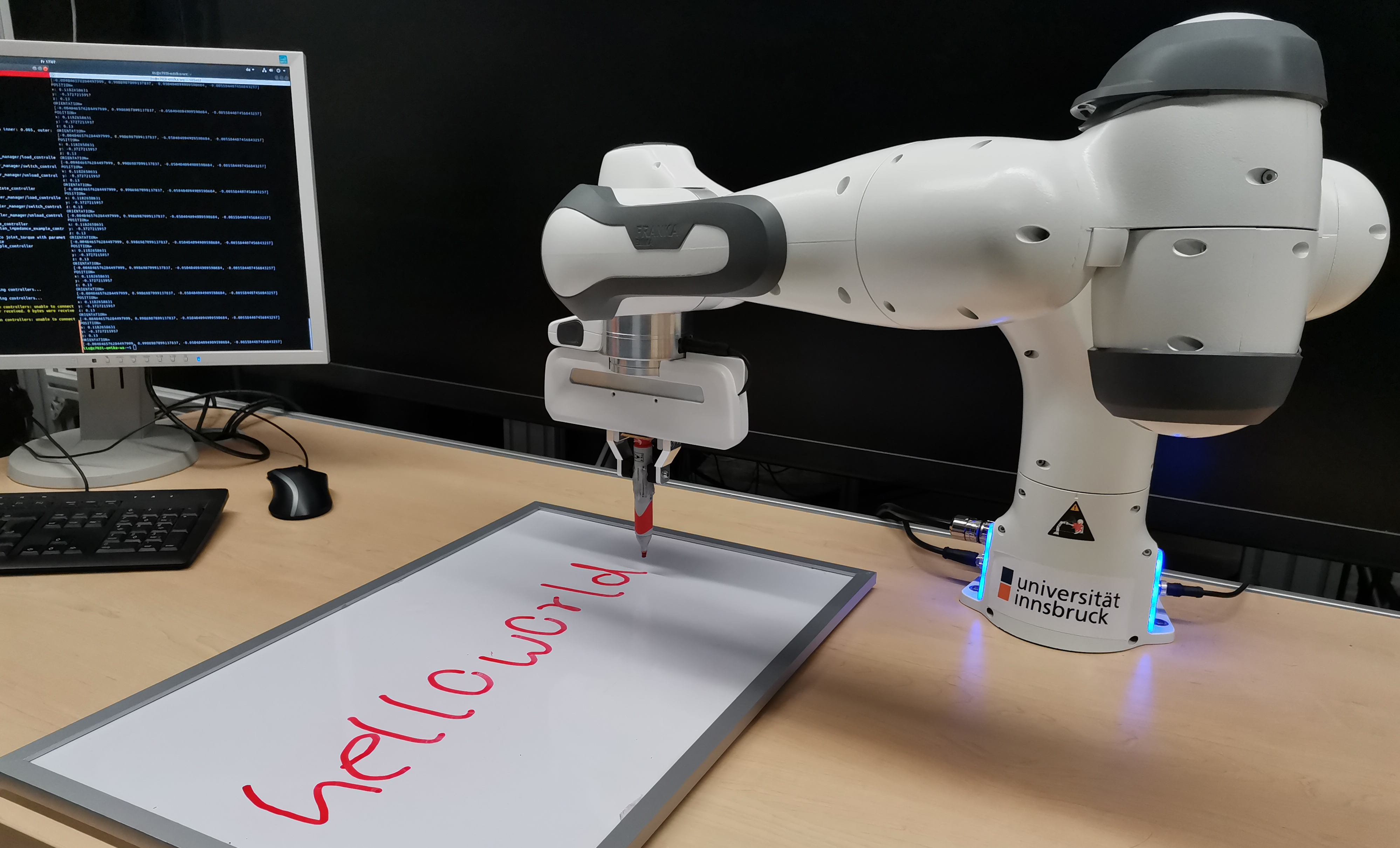}
	\caption{After being continually trained using learning from demonstration to write single letters, the robot can reproduce all the trajectories that it has learned in the past with a single HN network and without having access to training data from past tasks. Video is available at \url{https://youtu.be/0gdIImIBnXc}. 
	}
	\label{fig:robot}
\end{figure}

\begin{figure*}[t!]
\setcounter{figure}{13}
\centering
  \begin{tabular}{ccc}
  \subfloat[Position errors]{\includegraphics[width=0.48\textwidth]{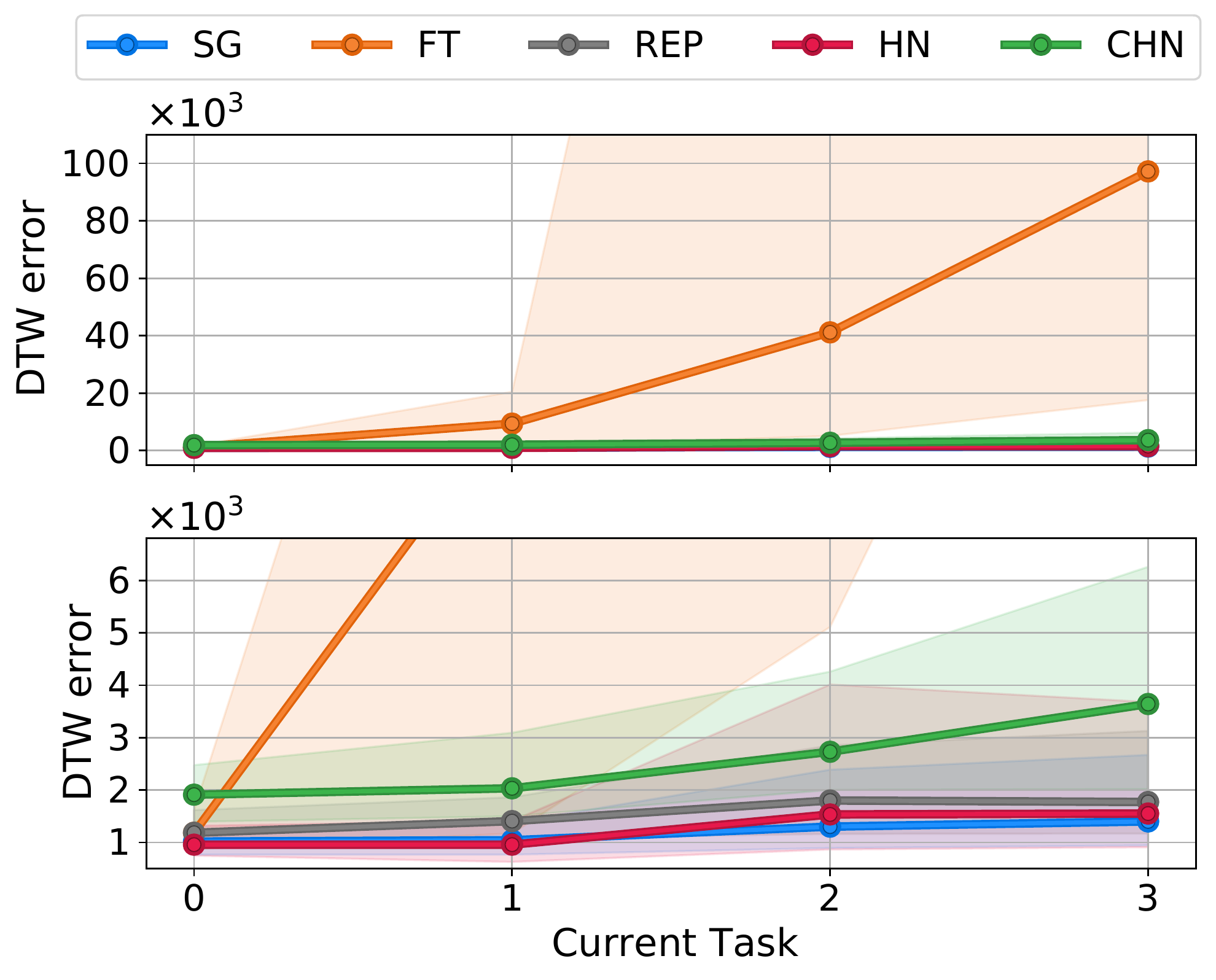}
  \label{fig:robottasks_cumu_pos}} &
  \subfloat[Orientation errors]{\includegraphics[width=0.48\textwidth]{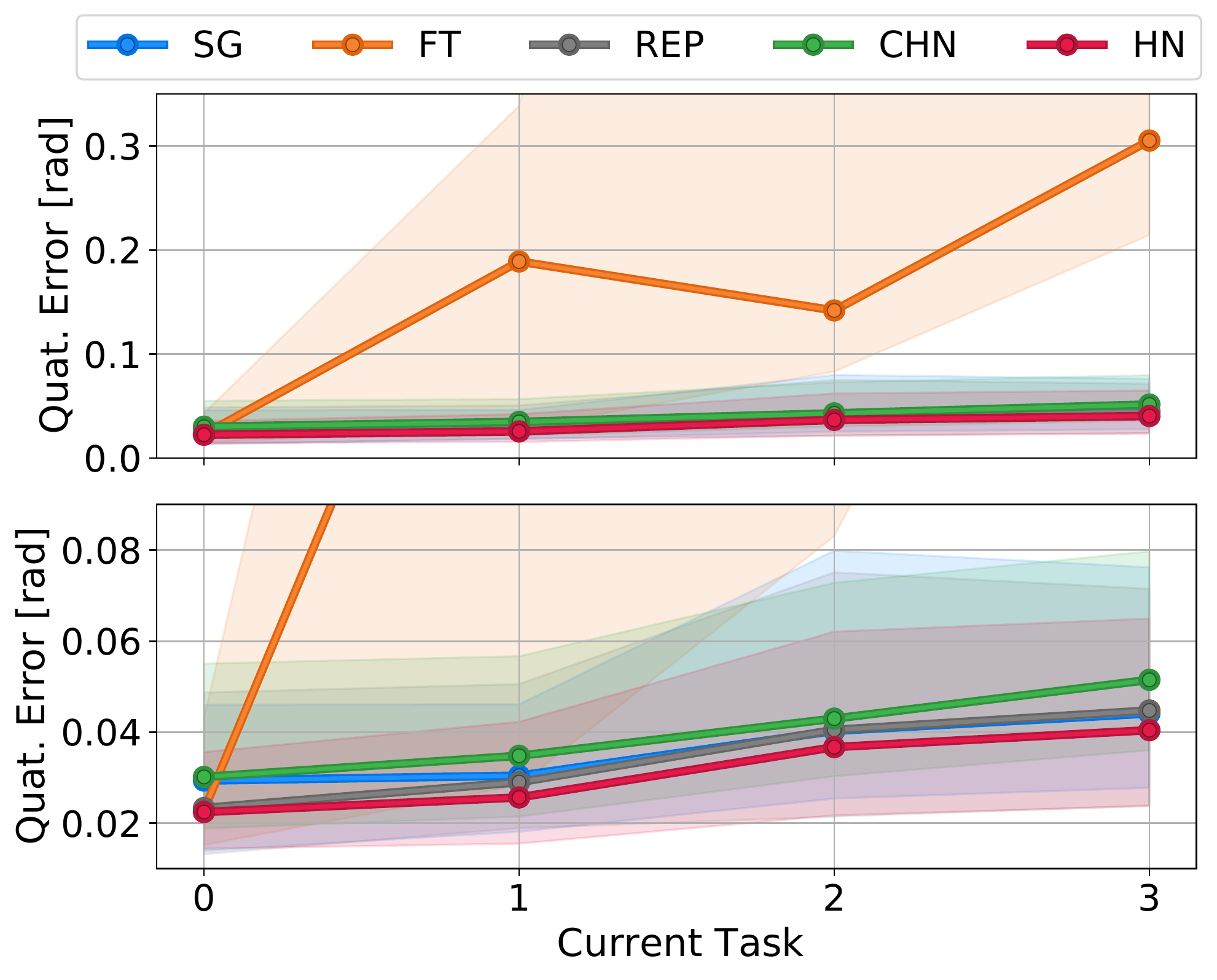}
  \label{fig:robottasks_cumu_ori}} &
  \end{tabular}
\caption[]{Position (a) and orientation (b) errors of trajectories predicted for the RoboTasks dataset (lower is better). The x-axis shows the \emph{current} task. After learning a task (using NODE$^\text{T}$), all current and previous tasks are evaluated. Lines show medians and shaded regions denote the lower and upper quartiles of the errors over 5 independent seeds. (top row) The errors for all methods are plotted. (bottom row) A detailed view of the methods that perform well. The performance of HN is comparable to the upper baseline SG.}
\label{fig:robottasks_cumu}
\end{figure*}

\begin{figure*}[b!]
	\setcounter{figure}{14}
	\includegraphics[width=\textwidth]{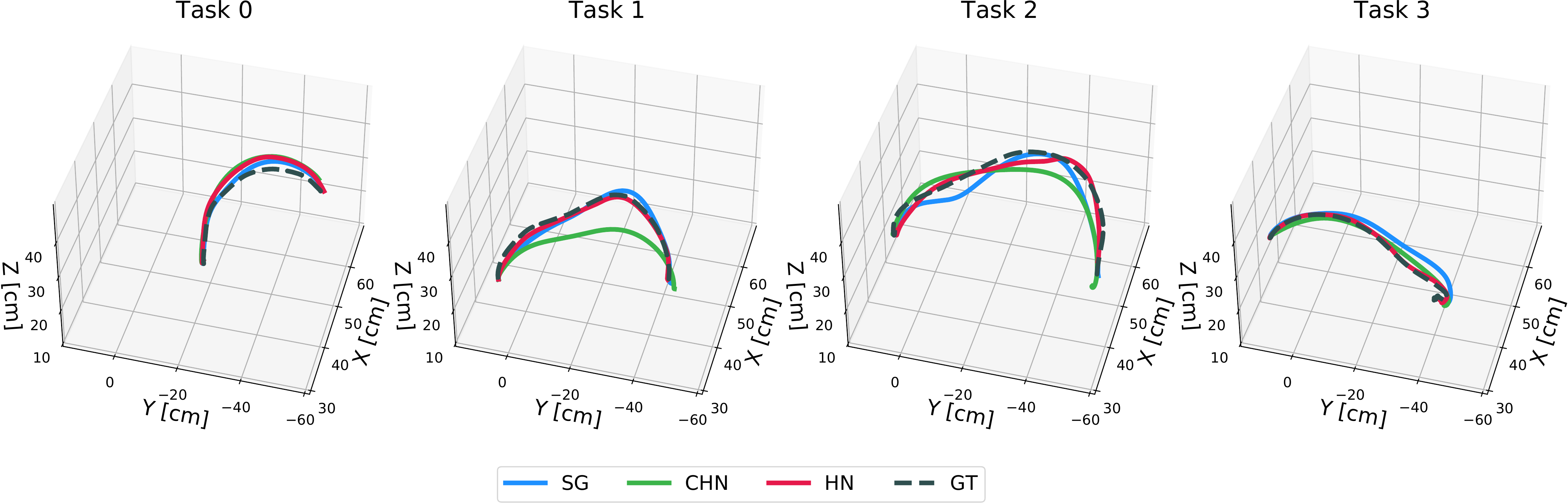}
	\caption{Example of end-effector position trajectories predicted for all past tasks after sequentially learning all the tasks of the RoboTasks dataset. All methods employ NODE$^\text{T}$ as the trajectory learning method. Hypernetworks (HN and CHN) remember all past tasks and mimic the demonstrations accurately.}
	\label{fig:robottasks_pos_traj}
\end{figure*}
\begin{figure*}[t!]
	\includegraphics[width=\textwidth]{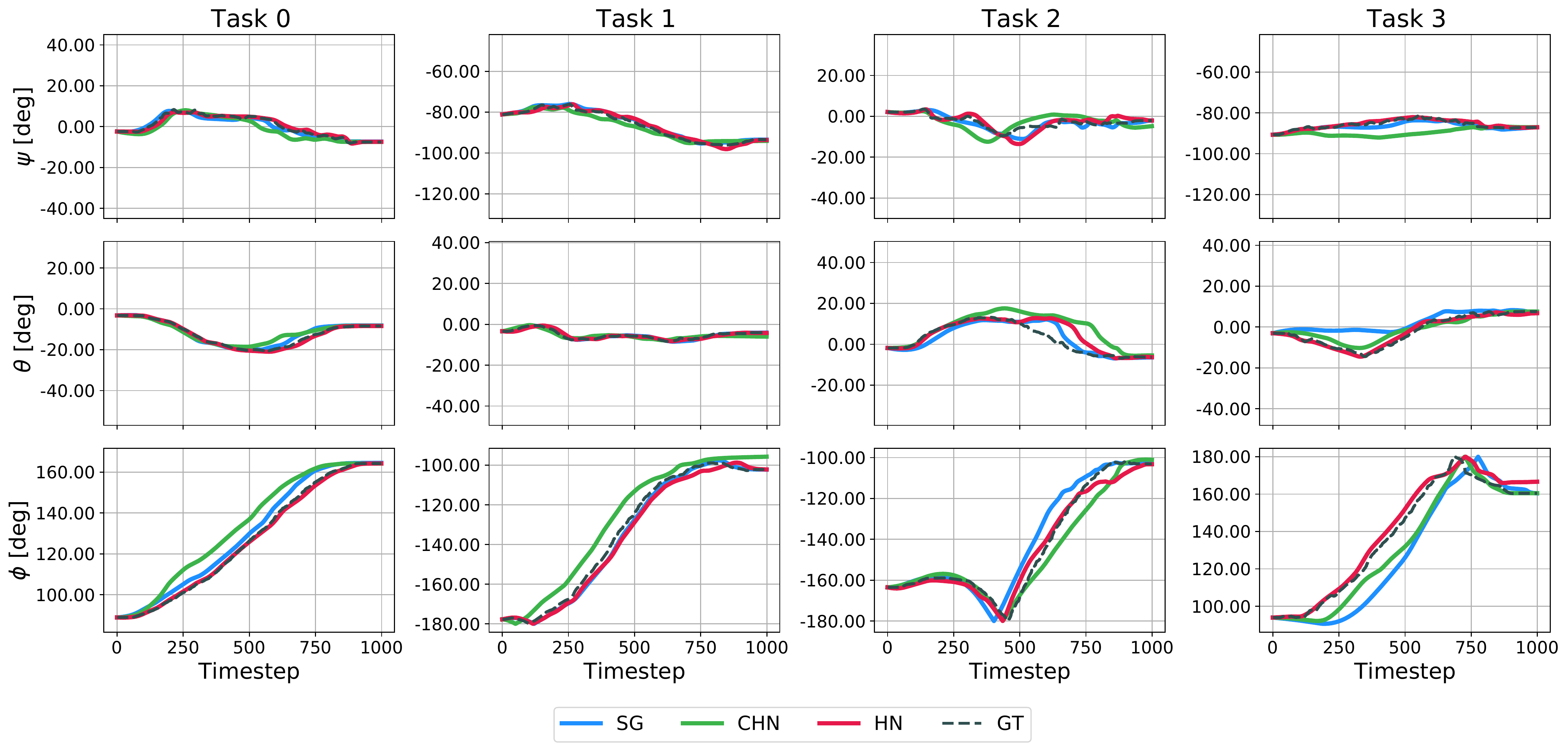}
	\caption{Example of end-effector orientation trajectories predicted for all past tasks after sequentially learning all the tasks of the RoboTasks dataset. All methods employ NODE$^\text{T}$ as the trajectory learning method. The ground truth demonstrations as well as the predictions of the orientation consist of a sequence of unit quaternions. Here, we convert each quaternion in the sequence to its corresponding Euler angle representation ($z$-$y$-$x$ convention) for easier interpretation. Columns show the predictions for each task, and rows show the different Euler angles in degrees. A figure with the original quaternion sequences can be seen in \myfigure{fig:robottasks_ori_traj_quat}. Hypernetworks (HN and CHN) remember all past tasks and mimic the ground truth orientations accurately.}
	\label{fig:robottasks_ori_traj_euler}
\end{figure*}

Using the same threshold computation approach we followed for $\bm{\mathcal{D}}_{\text{LASA}}$, we compute a DTW threshold value of 1821 for $\bm{\mathcal{D}}_{\text{HW}}$. With this, we compute the continual learning metrics shown in \mytable{tab:hw_cl}. The advantage of using NODEs with a time input is clear from the higher values of ACC for NODE$^{\text{T}}$ compared to NODE$^{\text{I}}$ for all the methods. In terms of ACC or REM, there is very little to choose between SG, REP, HN and CHN. However, when all the CL metrics are considered together (see the CL$_{sco}$ column in \mytable{tab:hw_cl}), CHN exhibits the best performance on account of its small size and the fact that it does not need to cache training data from past tasks. HN shows the second-best overall performance. 
\new{\myfigure{fig:spider_hw} illustrates that the hypernetwork methods (HN and CHN) perform well in all continual learning metrics, while the other methods perform well in only some aspects of continual learning but fail in others.}

We qualitatively evaluate how the trajectories predicted by HN can be reproduced with a physical robot. For this, we use the same Franka Emika Panda robot that was used for recording the demonstrations for $\bm{\mathcal{D}}_{\text{HW}}$. The HN model trained on the 7 tasks of $\bm{\mathcal{D}}_{\text{HW}}$ is queried to produce the letters \emph{h}, \emph{e}, \emph{l}, \emph{l}, \emph{o}, \emph{w}, \emph{o}, \emph{r}, \emph{l}, \emph{d} by using the appropriate task embedding vectors in sequence. The trajectory of each letter is scaled and translated by a constant amount and provided to the robot, which then follows this path with its end-effector. The $z$-coordinate and orientation of the end-effector are fixed. \myfigure{fig:robot} shows the letters written by the robot. 

\begin{figure}[t!]
	\includegraphics[width=\textwidth]{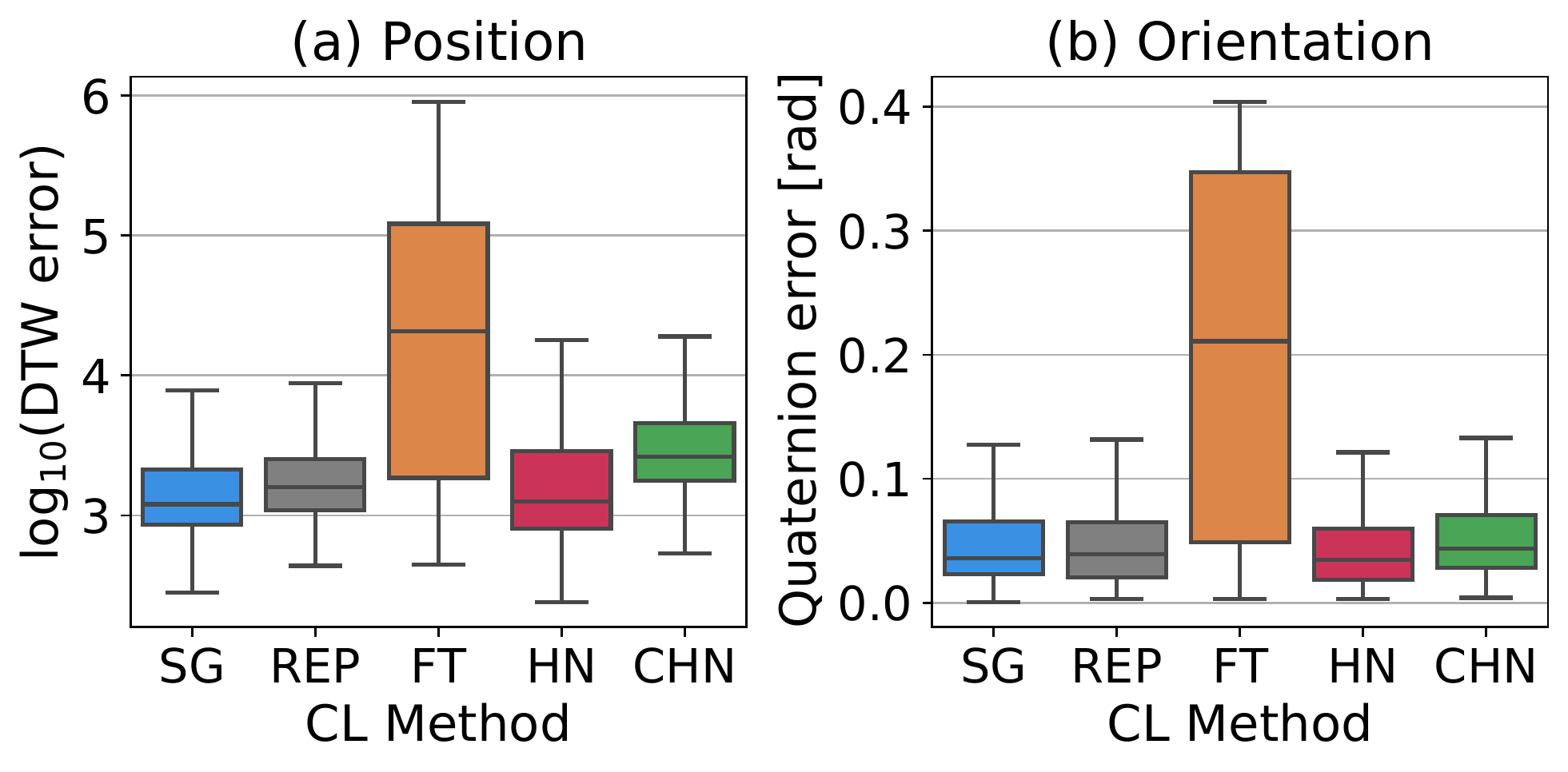}
	\caption{Prediction errors (lower is better) showing DTW position errors (a) and Quaternion orientation errors (b) of all trajectory predictions during the learning of all tasks of the RoboTasks dataset. Results are obtained using 5 independent seeds. Overall, HN and CHN produce low errors which are comparable to the performance of the upper baseline SG.}
	\label{fig:rt_pos_box}
\end{figure}


\subsubsection{RoboTasks}
\label{sec:res_robtasks}

$\bm{\mathcal{D}}_{\text{Robot}}$ comprises $4$ tasks, where each task involves learning the vector fields of the position as well as the orientation of the end-effector of the robot. For this experiment, the methods we evaluate are SG, FT, REP, CHN and HN. We omit SI and MAS because of their poor performance on the previous datasets. We use only NODE$^\text{T}$ for this experiment since in the previous experiments it has been shown to be better than NODE$^\text{I}$. As before, SG acts as an upper baseline and FT serves as a lower performance baseline. For each method, we train separate models for learning the position and orientation trajectories. The order of tasks for this experiment is: \emph{box opening}, \emph{bottle shelving}, \emph{plate stacking}, and \emph{pouring}.

\myfigure{fig:robottasks_cumu} shows the position and orientation errors of the predicted trajectories for all past and current tasks, as new tasks are learned. For both positions and orientations, FT is unable to predict correct trajectories for past tasks as soon as the second task is learned. The performance of the other methods is comparable to each other. For the end-effector position trajectories (\myfigure{fig:robottasks_cumu_pos}~(bottom)), the performance of SG, REP and HN are nearly identical, while CHN is slightly worse. However this difference does not result in significant qualitative differences in the robot's movement. SG, REP, CHN and HN are equally good at predicting the orientation trajectories, as can be seen in \myfigure{fig:robottasks_cumu_ori} (bottom). The slight upward trend in the plots of SG, REP, CHN and HN in \myfigure{fig:robottasks_cumu} can be attributed to the fact that task 2 (\emph{plate stacking}) is more difficult than the other tasks due to very diverse ground truth demonstrations.
\new{
It can be observed that CHN shows a more pronounced upward trend in the position error plot (\myfigure{fig:robottasks_cumu_pos} (bottom)). This is probably due to the fact that CHN has far fewer parameters than HN and thus has lower representation and remembrance capabilities.
}

Similar to the results for $\bm{\mathcal{D}}_{\text{LASA}}$ and $\bm{\mathcal{D}}_{\text{HW}}$, it can be seen that for continually learning 
\change{real-world}
robot tasks involving both position and orientation, the hypernetwork-based methods CHN and HN perform similar to SG and REP, but without needing significant growth in parameters and without requiring to be retrained on the data from prior tasks. 

\myfigure{fig:robottasks_pos_traj} shows an example of the position trajectories and \myfigure{fig:robottasks_ori_traj_euler} shows an example of the orientation trajectories predicted by SG, CHN and HN for tasks 0--4 after the last task has been learned (here we show the orientation in terms of Euler angles and the corresponding results in terms of quaternion elements can be seen in \myfigure{fig:robottasks_ori_traj_quat}). In both these qualitative examples, CHN and HN produce trajectories that closely mimic the ground truth demonstration. 
\myfigure{fig:rt_pos_box} shows the overall prediction errors during the course of training. Images of the robot performing the 4 tasks of $\bm{\mathcal{D}}_{\text{Robot}}$ after learning all the tasks sequentially with CHN can be seen in \myfigure{fig:robottasks_prediction}. 

\begin{table}[t!]
	\caption{Continual learning metrics for the RoboTasks dataset (median over 5 seeds). Values range from 0 (worst) to 1 (best).
	}
\subfloat[Position]{
	\centering
	\resizebox{\textwidth}{!}{
	\begin{tabular}{lllllllll}
\toprule
\textbf{MET} &   \textbf{ACC} &   \textbf{REM} &    \textbf{MS} &    \textbf{TE} &    \textbf{FS} &   \textbf{SSS} & \textbf{CL$_{sco}$} & \textbf{CL$_{stab}$} \\
\midrule
          SG &  \textbf{0.95} &  \textbf{1.00} &           0.52 &  \textbf{0.93} &           0.00 &  \textbf{1.00} &                0.73 &                 0.60 \\
          FT &           0.39 &           0.01 &  \textbf{1.00} &           0.84 &           0.69 &  \textbf{1.00} &                0.66 &                 0.61 \\
         REP &           0.94 &           0.94 &  \textbf{1.00} &           0.84 &           0.69 &           0.38 &                0.80 &                 0.77 \\
          HN &           0.88 &           0.98 &  \textbf{1.00} &           0.72 &           0.45 &  \textbf{1.00} &                0.84 &                 0.78 \\
         CHN &           0.90 &           0.90 &  \textbf{1.00} &           0.75 &  \textbf{0.74} &  \textbf{1.00} &       \textbf{0.88} &        \textbf{0.89} \\
\bottomrule
\end{tabular}

	}
	\label{tab:rt_pos_cl}
	}

	\subfloat[Orientation]{
	\centering
	\resizebox{\textwidth}{!}{
	\begin{tabular}{lllllllll}
\toprule
\textbf{MET} &   \textbf{ACC} &   \textbf{REM} &    \textbf{MS} &    \textbf{TE} &    \textbf{FS} &   \textbf{SSS} & \textbf{CL$_{sco}$} & \textbf{CL$_{stab}$} \\
\midrule
          SG &           0.98 &  \textbf{1.00} &           0.52 &           0.99 &           0.00 &  \textbf{1.00} &                0.75 &                 0.59 \\
          FT &           0.49 &           0.15 &  \textbf{1.00} &           0.99 &           0.66 &  \textbf{1.00} &                0.68 &                 0.62 \\
         REP &  \textbf{0.99} &  \textbf{1.00} &  \textbf{1.00} &  \textbf{1.00} &           0.66 &           0.38 &                0.84 &                 0.74 \\
          HN &  \textbf{0.99} &  \textbf{1.00} &  \textbf{1.00} &           0.77 &           0.20 &  \textbf{1.00} &                0.83 &                 0.68 \\
         CHN &           0.98 &           0.97 &  \textbf{1.00} &           0.78 &  \textbf{0.79} &  \textbf{1.00} &       \textbf{0.92} &        \textbf{0.90} \\
\bottomrule
\end{tabular}

	}
	\label{tab:rt_ori_cl}
	}
\label{tab:hw_rt}
\end{table}

Next, we report the continual learning metrics for $\bm{\mathcal{D}}_{\text{Robot}}$ in \mytable{tab:hw_rt}. For this we again need to set thresholds on the position and orientation errors to categorize predicted trajectories as accurate or inaccurate. This time we follow a slightly different approach for setting these thresholds compared to $\bm{\mathcal{D}}_{\text{LASA}}$ and $\bm{\mathcal{D}}_{\text{HW}}$. For positions, we first compute the DTW error between all pairs of demonstrations for each task. Since the ground truth demonstrations are already quite diverse, we use the maximum value from this list as the position error threshold. This threshold DTW value is 7131. For setting the orientation error threshold, we simply use a threshold of 10 degrees, since an absolute error of 10 degrees (averaged across the 3 rotation axes) results in roughly the same orientation as the desired one. 

\begin{figure}[t!]
	\centering
	\includegraphics[width=0.9\textwidth]{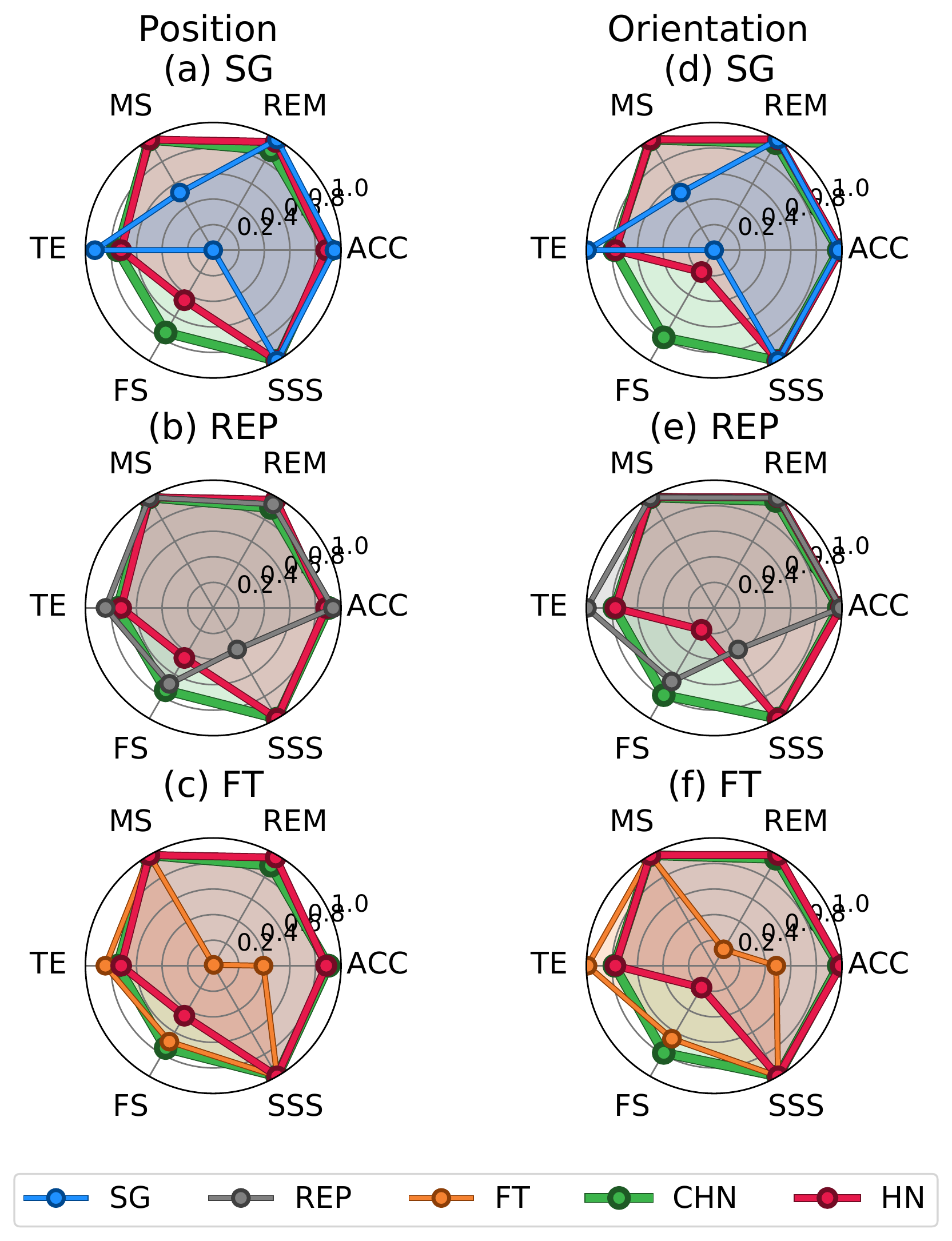}
	\caption{
	\new{Continual learning metrics (0: worst, 1: best) for the different methods (with NODE$^{\textrm{T}}$) on the RoboTasks dataset. In each individual plot, one of the baseline methods is compared against HN and CHN to illustrate the different tradeoffs made by the methods. The first column (a)-(c) shows the results for position, and the second column (d)-(f) shows the results for orientation. Across all metrics, CHN shows good performance for both position and orientation. HN exhibits a similar performance but due to its bigger size, achieves a low score for FS for orientation. REP's score for SSS is low due to its requirement of storing training data of all tasks. SG and FT achieve low scores in multiple metrics.
	}
	}
	\label{fig:spider_rt}
\end{figure}

For position as well as orientation, SG, REP, CHN and HN are comparable to each other in terms of ACC and REM. However, when all the metrics are considered together, CHN is the best amongst all the methods on account of its small near-constant parameter size and its non-dependence on the demonstrations of previous tasks. 
\new{
This is also corroborated by \myfigure{fig:spider_rt}, which shows that CHN performs well in all the continual learning metrics, while the other methods perform poorly in one or more metrics.
}

\balance 

\change{
We also evaluate the robustness of the predicted trajectories when the robot starts from novel initial positions that are not present in the demonstrations used for training. To this end, we compare the predictions made by upper-baseline SG and CHN for 100 randomly selected starting positions on each of the 4 tasks of the RoboTasks dataset. Please refer to \mysection{sec:appendix_robustness_start} in the appendix for details. Our results indicate that SG is much more sensitive to small changes in the starting position despite being the best at predicting trajectories that have the same starting position as the demonstrations. The continual learning process for CHN appears to reduce overfitting to the demonstrations and makes the model more robust to changes in the initial state of the robot. }

%
\section{Discussion}
\label{sec:discussion}

\change{
To be functional in the real world, a robot must be able to acquire multiple skills over time, without forgetting the skills that were learned in the past and also without needing to be retrained on past skills. In this paper, we have taken a further step in this direction by presenting the first work on continual trajectory learning from demonstrations.

The results from our experiments on 3 different datasets show that the hypernetwork model (HN) performs on par with the upper baseline (SG) for all datasets, and the smaller chunked hypernetwork (CHN) performs on par with SG on the robot datasets (HelloWorld and RoboTasks). Compared to the growth of SG for each new task, the hypernetwork models (HN, CHN) scale much more efficiently as the growth per task of HN and CHN is negligible (\mysection{sec:results}).

Using continual learning metrics (e.g. accuracy of predictions, how well past tasks are remembered, model size growth, storage of training data from past tasks, etc.), we compared the hypernetwork-based approach to methods from all continual learning families and found that empirically, hypernetworks perform best.
Specifically, HN is the best among the compared models for the LASA dataset.
On the remaining 2 robot datasets, the hypernetwork-based methods outperform the other approaches. CHN, due to its memory efficiency, is empirically the best, followed by HN. 

We also improve the trajectory learning NODE by introducing an additional time input to create the NODE$^\text{T}$ model. In \mysection{sec:res_hw}, we show how \mbox{NODE$^\text{T}$} improves performance and enables the models to learn trajectories with loops.

To facilitate future research in the area of continual-LfD, we release 2 new datasets of trajectories collected kinesthetically with a real robot: \emph{HelloWorld} (\mysection{sec:res_hw}) and \emph{RoboTasks} (\mysection{sec:res_robtasks}). The latter includes tasks involving changing positions and orientations. 
After continually learning all the tasks from these datasets using our hypernetwork-based approach, we verified that the real robot can successfully perform all past tasks.
Further, we verified that the order of tasks during training does not affect the performance of these models (refer to \mysection{sec:appendix_robustness_task_order} in the appendix for details). 

The code used in this paper is made publicly available to aid reproducibility.
Although the \mbox{HelloWorld} and \mbox{RoboTasks} datasets used in our experiments were recorded kinesthetically, our implementation only requires the training data in terms of the robot's trajectories and does not depend on how this data is collected. Similar to \citep{ahmadzadeh2018trajectory}, our implementation can potentially also be used for trajectories recorded using other means such as shadowing or teleoperation \citep{argall2009survey}.

A limitation of our current hypernetwork-based approach is the  increase in training time with an increasing number of tasks because an additional regularization term is added for each new task
(e.g. HN needs 620 seconds to learn task 0 and 1609 seconds to learn task 25 of the LASA dataset, further details in \mytable{tab:wall_clock}). As proposed by von Oswald et al.~\cite{von2019continual}, this can be overcome by 
using an approximate sample-based regularization instead of the full regularization process. 
We leave the experimental verification of this solution to future work.

In this paper, our focus is on continual learning and we use NODEs as a simple trajectory learning method because of its good empirical performance and fast convergence during training, which facilitates running multiple experiments on long sequences of tasks (\mysection{sec:met_lfd}). However, NODEs do not have a mechanism for enforcing the stability of the predicted trajectories. To overcome this, it is possible to substitute the simple NODE with a stable alternative in the future \cite{kolter2019learning, urain2020imitationflow}. 
Interestingly, we found that even without stability guarantees, NODEs generated by a chunked hypernetwork (CHN) produce trajectories that are much more robust to novel initial conditions than the upper-baseline SG, which uses separate networks to learn each individual task (refer to \mysection{sec:appendix_robustness_start} in the appendix for details). 

Further interesting research directions in continual LfD may include: improving the ability of the chunked hypernetworks to remember long sequences of tasks (e.g., via architectural changes), using continual reinforcement learning \cite{huang2021continual} to refine predicted trajectories \cite{calinon2013compliant}, and the inclusion of visual cues to improve and adapt trajectories~\cite{rahmatizadeh2018vision}. 

}

%

\section{Conclusion}
\label{sec:conclusion}

\change{
In this paper, we presented the first study on continual learning of trajectories from demonstrations, thereby expanding the field of continual learning in robotics to LfD. 
We adapted continual learning (CL) methods from all CL families, and compared these against our hypernetwork-based approaches on three different datasets.
Two of these three datasets were collected using kinesthetic teaching with the real robot and are released by us in this paper.
We have also verified that our approach works for real-world tasks involving changing positions and orientations by evaluating it on a real robot.
Compared to other continual learning approaches considered in this paper, hypernetworks achieve equal or better accuracy, but scale much better in terms of other criteria such as growth in model size, and storage of past training data.
Our results suggest that a regular hypernetwork (HN) is capable of learning many LfD tasks and is empirically the method of choice for a long sequence of tasks. For a shorter task sequence, the smaller chunked hypernetwork (CHN) is empirically the best choice for continual-LfD. 
}



%





\bibliographystyle{elsarticle-num} 
\biboptions{sort&compress}
\bibliography{references}


\clearpage

\appendix
\numberwithin{equation}{section}
\numberwithin{figure}{section}
\numberwithin{table}{section}

\renewcommand{\thesection}{\Alph{section}}

\section{Appendix}
\label{sec:appendix}

\subsection{Hardware Setup}
\label{sec:appendix_hardware}

We run all our experiments on a shared computing cluster with $125\,$GB of RAM and $16$ nodes, with each node having $1$ AMD Ryzen 2950X 16-Core processor and $4$ GeForce RTX 2070 GPUs. Each experiment uses a limited amount of RAM and only a single GPU and can be easily run on a single GPU or even on a CPU-only system.


\subsection{Training Times}
\label{sec:train_time}

\begin{table}[b!]
	\caption{Wall clock training times (in seconds) of the models for continual learning from demonstration (median over 5 independent seeds). }
\subfloat[LASA ($15 \times 10^3$ iterations/task)]{
	\centering
	\resizebox{\textwidth}{!}{
	\begin{tabular}{lrrrrrrr}
\toprule
METHOD &  Task 0 &  Task 5 &  Task 9 &  Task 13 &  Task 17 &  Task 21 &  Task 25 \\
\midrule
    SG &     586 &     691 &     714 &      728 &      677 &      721 &      737 \\
    FT &     665 &     722 &     767 &      752 &      673 &      747 &      735 \\
   REP &     792 &     909 &     900 &      913 &      904 &      934 &      936 \\
    SI &     694 &     699 &     707 &      702 &      647 &      713 &      686 \\
   MAS &     603 &     705 &     753 &      697 &      712 &      723 &      705 \\
    HN &     620 &    1087 &    1186 &     1253 &     1305 &     1510 &     1609 \\
   CHN &     604 &     991 &    1119 &     1257 &     1384 &     1624 &     1743 \\
\bottomrule
\end{tabular}

	}
	\label{tab:wall_clock_lasa}
	}

\subfloat[HelloWorld ($40 \times 10^3$ iterations/task)]{
	\centering
	\resizebox{\textwidth}{!}{
	\begin{tabular}{lrrrrrrr}
\toprule
METHOD &  Task 0 &  Task 1 &  Task 2 &  Task 3 &  Task 4 &  Task 5 &  Task 6 \\
\midrule
    SG &    1896 &    2012 &    1872 &    1960 &    2030 &    2030 &    2281 \\
    FT &    1954 &    2022 &    1973 &    2016 &    2050 &    2086 &    2385 \\
   REP &    1926 &    1939 &    1950 &    1942 &    1957 &    1975 &    2016 \\
    SI &    2019 &    2128 &    2093 &    2091 &    2110 &    2132 &    2268 \\
   MAS &    1965 &    2281 &    2252 &    2302 &    2316 &    2362 &    2500 \\
    HN &    1968 &    2337 &    2508 &    2702 &    2843 &    2930 &    3234 \\
   CHN &    1948 &    2254 &    2341 &    2578 &    2786 &    2858 &    3150 \\
\bottomrule
\end{tabular}

	}
	\label{tab:wall_clock_hw}
 	}

\subfloat[RoboTasks-position ($50 \times 10^3$ iterations/task)]{
	\centering
	\resizebox{0.68\textwidth}{!}{
	\begin{tabular}{lrrrr}
\toprule
METHOD &  Task 0 &  Task 1 &  Task 2 &  Task 3 \\
\midrule
    SG &    1869 &    2400 &    2328 &    1684 \\
    FT &    1483 &    1985 &    1923 &    1732 \\
   REP &    1486 &    1952 &    1901 &    1797 \\
    HN &    1508 &    2387 &    2585 &    2214 \\
   CHN &    1591 &    2484 &    2662 &    2141 \\
\bottomrule
\end{tabular}

	}
	\label{tab:wall_clock_rtpos}
	}

\subfloat[RoboTasks-orientation ($50 \times 10^3$ iterations/task)]{
	\centering
	\resizebox{0.68\textwidth}{!}{
	\begin{tabular}{lrrrr}
\toprule
METHOD &  Task 0 &  Task 1 &  Task 2 &  Task 3 \\
\midrule
    SG &    1809 &    1800 &    1800 &    1783 \\
    FT &    2026 &    2068 &    2061 &    2058 \\
   REP &    1396 &    1409 &    1409 &    1409 \\
    HN &    1461 &    1903 &    2116 &    2303 \\
   CHN &    1504 &    1840 &    2204 &    2511 \\
\bottomrule
\end{tabular}

	}
	\label{tab:wall_clock_rtori}
 	}	

\label{tab:wall_clock}
\end{table}

In \mytable{tab:wall_clock}, we show the time (in seconds) needed to train each of the 7 continual learning methods presented in this paper (median values over 5 independent seeds). The hypernetwork-based methods (HN and CHN) show a small gradual increase in the training time as the number of tasks increase. This is due to the extra regularization that needs to be performed to prevent catastrophic forgetting. A possible way to attain a near constant training time (as proposed by von Oswald et al.~\cite{von2019continual}) is to randomly sample a fixed set of task embedding vectors to regularize with instead of using the embeddings of all tasks in each training iteration. We do not apply this process in this paper, but leave it for future work.


\subsection{Inference Times}
\label{sec:inference_time}

\begin{table}[b!]
	\caption{Inference speeds of trained hypernetworks (HN and CHN) in milliseconds and Hz (median over 100 forward passes).
	$\bold{f}_{\mathbf{h}} \rightarrow \bold{f}_{\bm{\uptheta}}$ denotes the generation of a NODE by a hypernetwork,
	$\bold{f}_{\bm{\uptheta}}(\mathbf{x})=\dot{\mathbf{x}}$ denotes a single step taken by the generated NODE, and
	$\int\bold{f}_{\bm{\uptheta}}(\mathbf{x}_\tau)\mathrm{d}\tau$ denotes the generation of an entire trajectory from a NODE by integrating over 1000 steps.
	}
\subfloat[GPU]{
	\centering
	\resizebox{\textwidth}{!}{
	\begin{tabular}{llrrrrrr}
\toprule
    Dataset      & Method & \multicolumn{2}{c}{$\bold{f}_{\mathbf{h}} \rightarrow \bold{f}_{\bm{\uptheta}}$} & \multicolumn{2}{c}{$\bold{f}_{\bm{\uptheta}}(\mathbf{x})=\dot{\mathbf{x}}$} & \multicolumn{2}{c}{$\int\bold{f}_{\bm{\uptheta}}(\mathbf{x}_\tau)\mathrm{d}\tau$} \\
                 &        &                             (ms)                             &   (Hz)  &                           (ms)                          &   (Hz)  &                              (ms)                             & (Hz) \\
\midrule
            LASA &    HN  &                        0.73                        & 1373.83 &                        0.27                        & 3765.08 &                       584.27                       & 1.71 \\
            LASA &   CHN  &                        0.61                        & 1642.25 &                        0.27                        & 3771.86 &                       569.71                       & 1.76 \\
      HelloWorld &    HN  &                        0.72                        & 1391.15 &                        0.26                        & 3856.83 &                       562.79                       & 1.78 \\
      HelloWorld &   CHN  &                        0.60                        & 1662.43 &                        0.27                        & 3731.59 &                       569.51                       & 1.76 \\
 RoboTasks (pos) &    HN  &                        0.74                        & 1360.24 &                        0.27                        & 3761.71 &                       570.53                       & 1.75 \\
 RoboTasks (pos) &   CHN  &                        0.60                        & 1660.45 &                        0.27                        & 3728.27 &                       570.28                       & 1.75 \\
 RoboTasks (ori) &    HN  &                        0.74                        & 1348.00 &                        0.27                        & 3685.68 &                       595.44                       & 1.68 \\
 RoboTasks (ori) &   CHN  &                        0.63                        & 1597.83 &                        0.28                        & 3548.48 &                       574.78                       & 1.74 \\
\bottomrule
\end{tabular}

	}
	\label{tab:inference_gpu}
	}
 
\subfloat[CPU]{
	\centering
	\resizebox{\textwidth}{!}{
	\begin{tabular}{llrrrrrr}
\toprule
    Dataset      & Method & \multicolumn{2}{c}{$\bold{f}_{\mathbf{h}} \rightarrow \bold{f}_{\bm{\uptheta}}$} & \multicolumn{2}{c}{$\bold{f}_{\bm{\uptheta}}(\mathbf{x})=\dot{\mathbf{x}}$} & \multicolumn{2}{c}{$\int\bold{f}_{\bm{\uptheta}}(\mathbf{x}_\tau)\mathrm{d}\tau$} \\
                 &        &                             (ms)                             &  (Hz)  &                           (ms)                          &   (Hz)  &                              (ms)                             & (Hz) \\
\midrule
            LASA &    HN  &                        3.01                        & 332.56 &                        0.12                        & 8422.30 &                       261.13                       & 3.83 \\
            LASA &   CHN  &                        5.23                        & 191.04 &                        0.61                        & 1634.89 &                       730.18                       & 1.37 \\
      HelloWorld &    HN  &                        2.94                        & 339.99 &                        0.12                        & 8456.26 &                       262.61                       & 3.81 \\
      HelloWorld &   CHN  &                        5.29                        & 189.14 &                        0.66                        & 1511.73 &                       750.89                       & 1.33 \\
 RoboTasks (pos) &    HN  &                        2.87                        & 348.16 &                        0.17                        & 5833.52 &                       257.85                       & 3.88 \\
 RoboTasks (pos) &   CHN  &                        5.78                        & 172.94 &                        0.57                        & 1749.81 &                       738.15                       & 1.35 \\
 RoboTasks (ori) &    HN  &                        7.51                        & 133.10 &                        0.15                        & 6579.30 &                       265.23                       & 3.77 \\
 RoboTasks (ori) &   CHN  &                       17.68                        &  56.55 &                        1.22                        &  820.88 &                      1499.81                       & 0.67 \\
\bottomrule
\end{tabular}

	}
	\label{tab:inference_cpu}
 	}	

\label{tab:inference}
\end{table}

\begin{table*}[b!]
	\caption{Hyperparameters for the LASA dataset. }
	\centering
	\resizebox{0.68\textwidth}{!}{
	\begin{tabular}{llllllll}
\toprule
    Hyperparameter &              SG &              FT &             REP &              SI &             MAS &             HN &             CHN \\
\midrule
  Train iterations &           15000 &           15000 &           15000 &           15000 &           15000 &          15000 &           15000 \\
     Learning rate &          0.0001 &          0.0001 &          0.0001 &          0.0001 &          0.0001 &         0.0001 &          0.0001 \\
  NODE output dim. &               2 &               2 &               2 &               2 &               2 &              2 &               2 \\
NODE hidden layers & [1000]$\times$3 & [1000]$\times$3 & [1000]$\times$3 & [1000]$\times$3 & [1000]$\times$3 & [100]$\times$3 & [1000]$\times$3 \\
   NODE activation &             elu &             elu &             elu &             elu &             elu &            elu &             elu \\
    Task Emb. dim. &               - &             256 &             256 &             256 &             256 &            256 &             256 \\
            c (SI) &               - &               - &               - &             0.3 &               - &              - &               - \\
        $\xi$ (SI) &               - &               - &               - &             0.3 &               - &              - &               - \\
   $\lambda$ (MAS) &               - &               - &               - &               - &             0.1 &              - &               - \\
  HN hidden layers &               - &               - &               - &               - &               - & [200]$\times$3 &  [200]$\times$3 \\
      $\beta$ (HN) &               - &               - &               - &               - &               - &          0.005 &           0.005 \\
   Chunk Emb. dim. &               - &               - &               - &               - &               - &              - &             256 \\
        Chunk dim. &               - &               - &               - &               - &               - &              - &            8192 \\
\bottomrule
\end{tabular}

	\label{tab:hparam_lasa}
	}
\end{table*}

\begin{table*}[b!]
	\caption{Hyperparameters for the HelloWorld dataset. }
	\centering
	\resizebox{0.68\textwidth}{!}{
	\begin{tabular}{llllllll}
\toprule
    Hyperparameter &              SG &              FT &             REP &              SI &             MAS &             HN &             CHN \\
\midrule
  Train iterations &           40000 &           40000 &           40000 &           40000 &           40000 &          40000 &           40000 \\
     Learning rate &          0.0001 &          0.0001 &          0.0001 &          0.0001 &          0.0001 &         0.0001 &          0.0001 \\
  NODE output dim. &               2 &               2 &               2 &               2 &               2 &              2 &               2 \\
NODE hidden layers & [1000]$\times$3 & [1000]$\times$3 & [1000]$\times$3 & [1000]$\times$3 & [1000]$\times$3 & [100]$\times$3 & [1000]$\times$3 \\
   NODE activation &             elu &             elu &             elu &             elu &             elu &            elu &             elu \\
    Task Emb. dim. &               - &             256 &             256 &             256 &             256 &            256 &             256 \\
            c (SI) &               - &               - &               - &             0.3 &               - &              - &               - \\
        $\xi$ (SI) &               - &               - &               - &             0.3 &               - &              - &               - \\
   $\lambda$ (MAS) &               - &               - &               - &               - &             0.1 &              - &               - \\
  HN hidden layers &               - &               - &               - &               - &               - & [200]$\times$3 &  [200]$\times$3 \\
      $\beta$ (HN) &               - &               - &               - &               - &               - &          0.005 &           0.005 \\
   Chunk Emb. dim. &               - &               - &               - &               - &               - &              - &             256 \\
        Chunk dim. &               - &               - &               - &               - &               - &              - &            8192 \\
\bottomrule
\end{tabular}

	\label{tab:hparam_hw}
	}
\end{table*}

The time taken by a network to produce predictions is critically important in a robotics scenario. The hypernetwork-based solutions for \emph{learning from demonstration} are well suited for this task, as can be seen by the speed of making predictions reported in \mytable{tab:inference}. For this evaluation, the models which were originally trained on the GPU, were loaded either into GPU memory (\mytable{tab:inference} (a)) or into the CPU memory (\mytable{tab:inference} (b)). After loading the hypernetworks (HN or CHN), one of the task embedding vectors was used as an input to \emph{generate} the parameters for a NODE and the speed of this forward pass is noted (column $\bold{f}_{\mathbf{h}} \rightarrow \bold{f}_{\bm{\uptheta}}$ in \mytable{tab:inference} (a) and (b)). 
Note that this NODE generation step occurs only when the robot is required to perform a new task (task switch). 
Once a NODE for a task is generated, it can be reused for multiple predictions of that task. 
The generated NODE is then made to predict a trajectory and the speed of each step (column $\bold{f}_{\bm{\uptheta}}(\mathbf{x})=\dot{\mathbf{x}}$), as well as the time to integrate over a 1000 steps (column $\int\bold{f}_{\bm{\uptheta}}(\mathbf{x}_\tau)\mathrm{d}\tau$) is noted. 
This process is repeated 100 times and the median times and speeds are reported in milliseconds and Hz. As expected, using the GPU generally results in faster predictions. 
The time taken to integrate over the entire trajectory of 1000 steps is more than a thousand times the time for a single step (due to the integration operation) and in the majority of cases, this takes much less than 1 second.
It can be seen from \mytable{tab:inference} that hypernetworks are able to produce predictions very fast on both the GPU as well as the CPU. This makes them suitable for use in a robotics application.
%

\subsection{Hyperparameters}
\label{sec:appendix_hyperparameters}

The hyperparameters used for the LASA, HelloWorld and RoboTasks datasets are listed in Tables \ref{tab:hparam_lasa}, \ref{tab:hparam_hw} and \ref{tab:hparam_rt} respectively. The regularization hyperparameters for SI ($c, \xi$), MAS ($\lambda$) and HN/CHN ($\beta$) are based on values proposed by \cite{zenke2017SI}, \cite{aljundi2018MAS} and \cite{von2019continual} respectively. Earlier, in \mysection{sec:results} we have shown that HN and CHN are robust to changes in regularization hyperparameters. In all cases, NODEs use the smooth ELU activation function, and hypernetworks have ReLU activations. We use the Adam optimizer in all our experiments.

\begin{table*}[t!]
	\caption{Hyperparameters for the RoboTasks dataset. }
\subfloat[Position]{
	\centering
	\resizebox{0.48\textwidth}{!}{
	\begin{tabular}{llllll}
\toprule
    Hyperparameter &              SG &              FT &             REP &             HN &             CHN \\
\midrule
  Train iterations &           50000 &           50000 &           50000 &          50000 &           50000 \\
     Learning rate &         0.00001 &         0.00001 &         0.00001 &        0.00001 &         0.00001 \\
  NODE output dim. &               3 &               3 &               3 &              3 &               3 \\
NODE hidden layers & [1000]$\times$3 & [1000]$\times$3 & [1000]$\times$3 & [100]$\times$3 & [1000]$\times$3 \\
   NODE activation &             elu &             elu &             elu &            elu &             elu \\
    Task Emb. dim. &               - &             512 &             512 &            512 &             512 \\
  HN hidden layers &               - &               - &               - & [200]$\times$3 &  [200]$\times$3 \\
      $\beta$ (HN) &               - &               - &               - &         0.0005 &          0.0005 \\
   Chunk Emb. dim. &               - &               - &               - &              - &             512 \\
        Chunk dim. &               - &               - &               - &              - &            8192 \\
\bottomrule
\end{tabular}

	}
	\label{tab:hparam_rtpos}
 	}	
\subfloat[Orientation]{
	\centering
	\resizebox{0.52\textwidth}{!}{
	\begin{tabular}{llllll}
\toprule
    Hyperparameter &              SG &              FT &             REP &          HN &             CHN \\
\midrule
  Train iterations &           50000 &           50000 &           50000 &       50000 &           50000 \\
     Learning rate &          0.0001 &          0.0001 &          0.0001 &      0.0001 &          0.0001 \\
  NODE output dim. &               3 &               3 &               3 &           3 &               3 \\
NODE hidden layers & [1500]$\times$3 & [1500]$\times$3 & [1500]$\times$3 & 150,150,150 & [1500]$\times$3 \\
   NODE activation &             elu &             elu &             elu &         elu &             elu \\
    Task Emb. dim. &               - &            1024 &            1024 &        1024 &            1024 \\
  HN hidden layers &               - &               - &               - & 300,300,300 &     300,300,300 \\
      $\beta$ (HN) &               - &               - &               - &      0.0005 &          0.0005 \\
   Chunk Emb. dim. &               - &               - &               - &           - &            1024 \\
        Chunk dim. &               - &               - &               - &           - &            8192 \\
\bottomrule
\end{tabular}

	}
	\label{tab:tab:hparam_rtori}
 	}	
\label{tab:hparam_rt}
\end{table*}

\begin{figure*}[b!]
	\setcounter{figure}{1}
	\includegraphics[width=\textwidth]{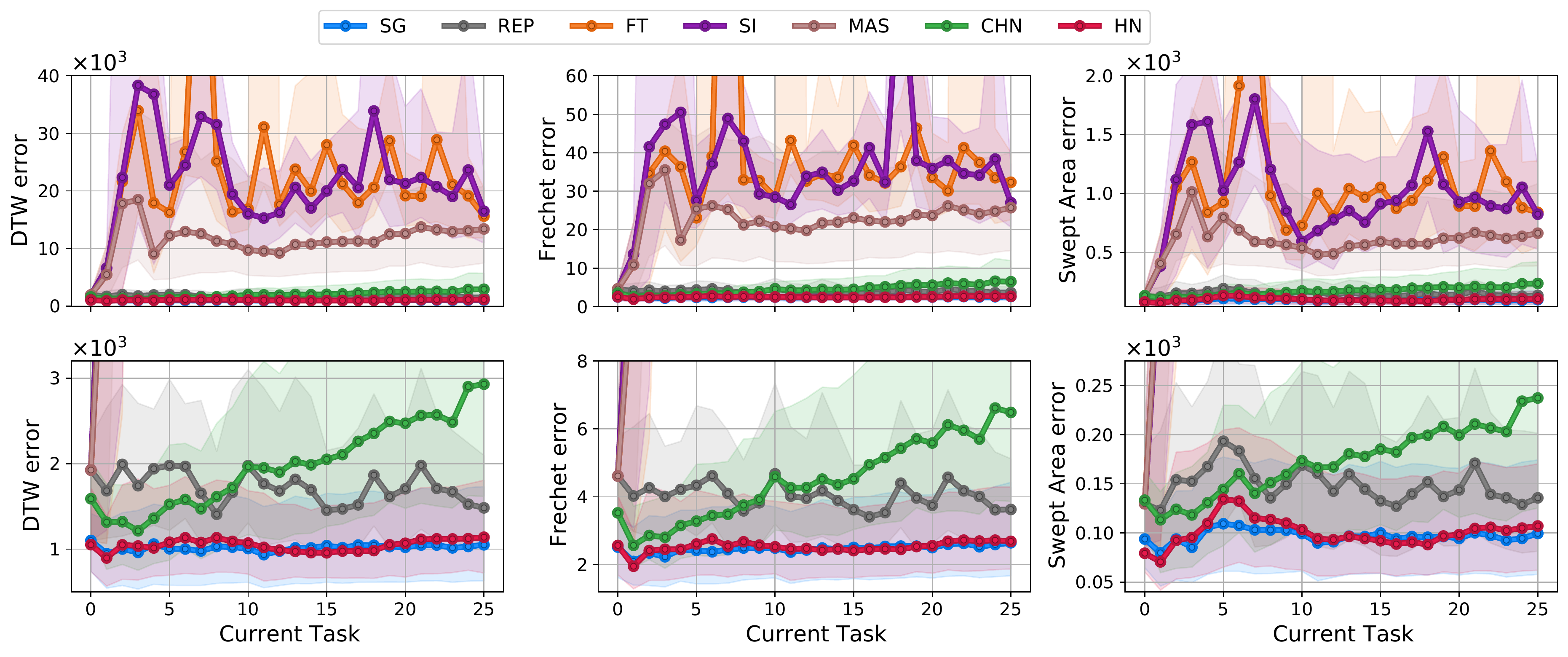}
	\caption{DTW, Frechet distance and Swept Area errors of trajectories predicted for the LASA dataset (lower is better). The x-axis shows the \emph{current} task. After learning a task (using NODE$^\text{T}$), all current and previous tasks are evaluated. Plots for SG and HN overlap with each other. Lines show medians and shaded regions denote the lower and upper quartiles of the errors over 5 independent seeds. (top) The errors for all methods. (bottom) A zoomed-in view of the methods that perform well.}
	\label{fig:lasa_cumu_all_metrics}
\end{figure*}

\begin{figure}[b!]
	\setcounter{figure}{0}
    \centering
    $\vcenter{
    \hfill
    \includegraphics[width=0.43\textwidth]{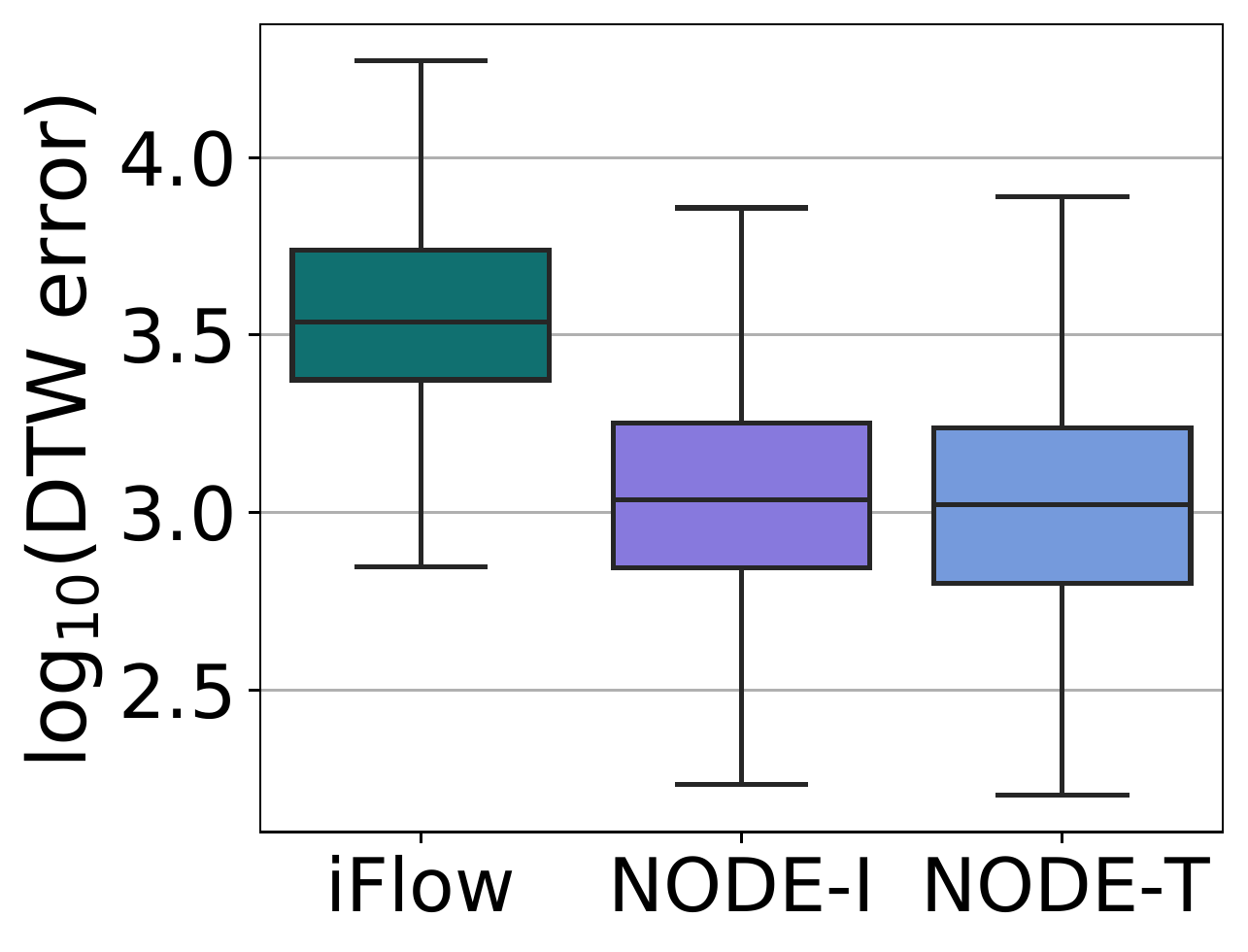}
    \hfill
    \resizebox{0.45\textwidth}{!}{
    \begin{tabular}[b]{lr}\hline
      Hyperparameter    &     iFlow \\ \hline
      Batch size        &       256 \\
      Depth             &         7 \\
      Learning rate     &    0.0001 \\
      Epochs            &      1000 \\
      Train iterations  &     27000 \\
      Output dim.       &         2 \\ \hline
    \end{tabular}
    }
    \hfill
    }$
    \caption{Comparison of NODE with iFlow. (left) DTW errors (lower is better) of all trajectory predictions for all tasks of the LASA dataset. Standalone models of iFlow, NODE$^\text{T}$, and NODE$^\text{I}$ are used to learn each of the 26 LASA tasks. Results are obtained using 5 independent seeds. (right) Hyperparameters for iFlow. Hyperparameters for NODE are given in \mytable{tab:hparam_lasa}. NODEs show better empirical performance and converge faster than iFlow.}
    \label{fig:iflow}
\end{figure}

\begin{figure*}[b!]
	\setcounter{figure}{2}
	\includegraphics[width=\textwidth]{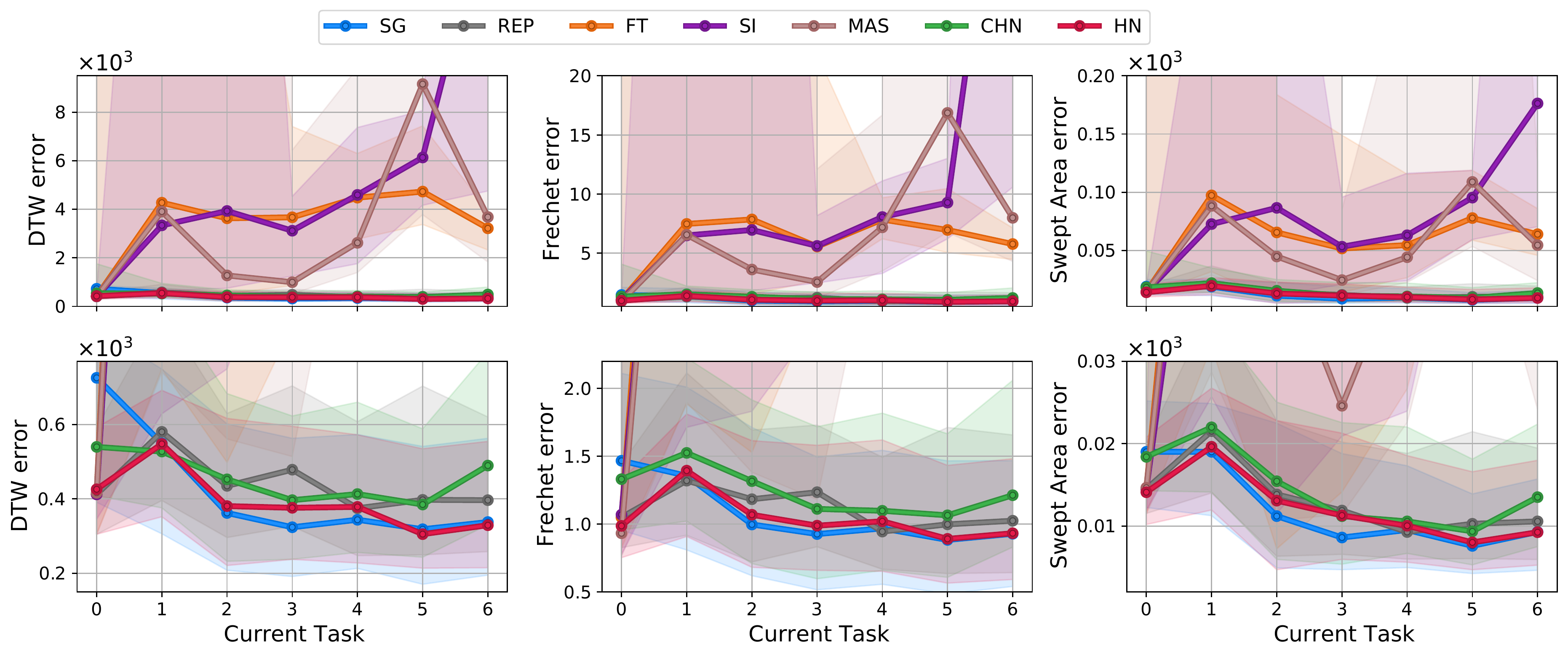}
	\caption{DTW, Frechet distance and Swept Area errors of trajectories predicted for the HelloWorld dataset (lower is better). The x-axis shows the \emph{current} task. After learning a task (using NODE$^\text{T}$), all current and previous tasks are evaluated. Lines show medians and shaded regions denote the lower and upper quartiles of the errors over 5 independent seeds. (top) The errors for all methods. (bottom) A zoomed-in view of the methods that perform well.}
	\label{fig:hw_cumu_all_metrics}
\end{figure*}

\begin{figure*}[b!]
	\includegraphics[width=0.98\textwidth]{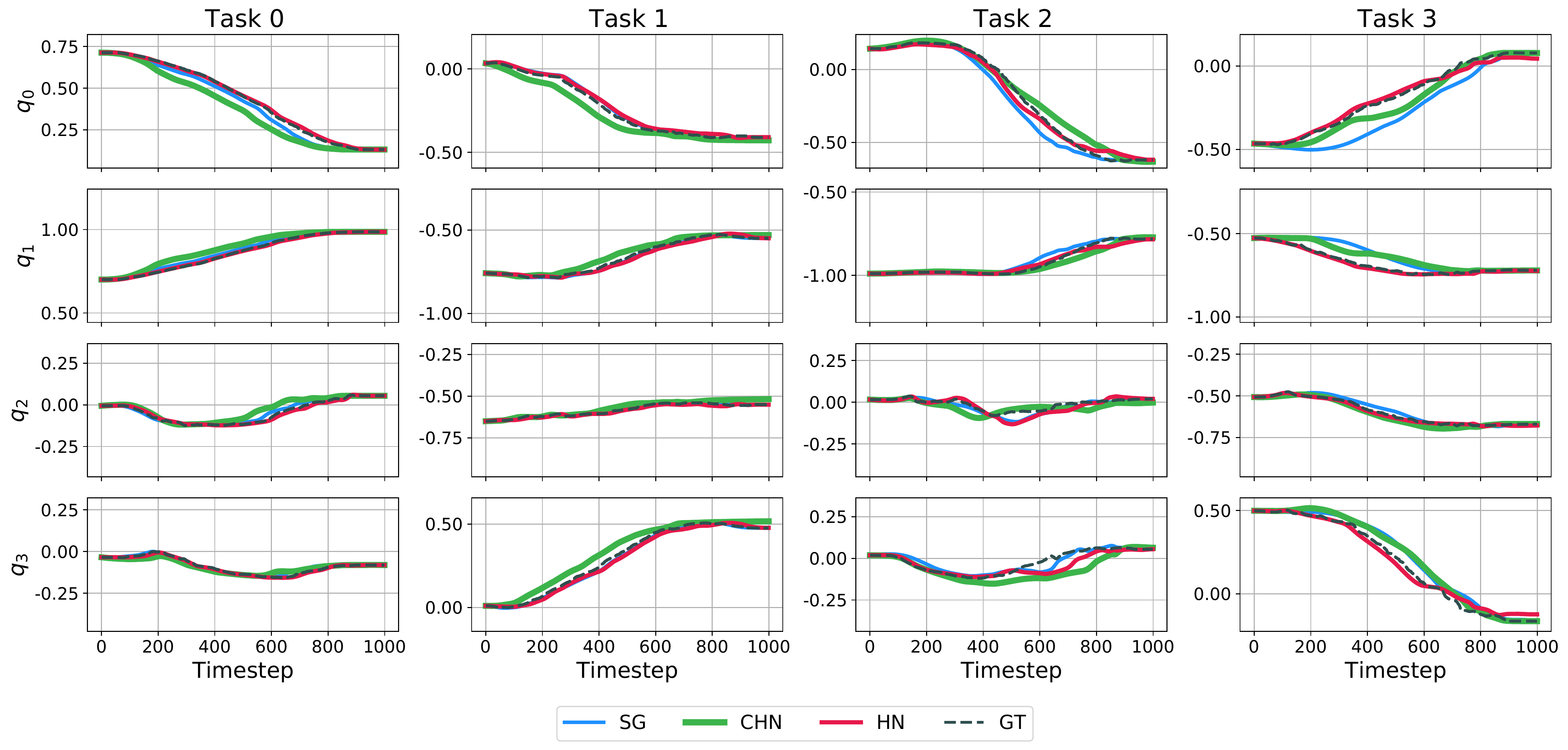}
	\caption{Example of end-effector orientation trajectories predicted for all past tasks after sequentially learning all the tasks of the RoboTasks dataset. All methods employ NODE$^\text{T}$ as the trajectory learning method. The ground truth demonstrations as well as the predictions of the orientation consist of a sequence of unit quaternions. Columns show the predictions for each task, and rows show the different elements of the quaternions (scalar first convention).}
	\label{fig:robottasks_ori_traj_quat}
\end{figure*}
\vfill


\subsection{Imitation Flow Comparison}
\label{sec:iflow}

To compare NODEs against an alternative trajectory learning approach, we performed a small non-continual learning experiment using the LASA dataset. For each task, we trained a model of iFlow~\citep{urain2020imitationflow}, NODE$^\text{I}$, and NODE$^\text{T}$ and measured the DTW errors (\myfigure{fig:iflow} shows the errors for all tasks taken together). 
The iFlow hyperparameters are shown in \myfigure{fig:iflow} (right) and the NODE hyperparameters are given in \mytable{tab:hparam_lasa}.
Although for iFlow we used almost twice the number of training iterations as NODEs, the empirical performance of NODEs is still better. 
An iFlow model needs around 60 minutes to be trained to convergence for a single LASA task on our setup, whereas a NODE takes around 10 minutes. 
 
\subsection{Additional Results}
\label{sec:appendix_additional_res}

Figs. \ref{fig:lasa_cumu_all_metrics} and \ref{fig:hw_cumu_all_metrics} report additional trajectory error metrics (\emph{Frechet distance} and \emph{Swept Area error}) for the LASA and HelloWorld datasets respectively. \myfigure{fig:robottasks_ori_traj_quat} shows a qualitative example of orientation prediction in terms of quaternions for the RoboTasks dataset.

\begin{figure*}[tb]
    \centering
    \includegraphics[width=0.93\textwidth]{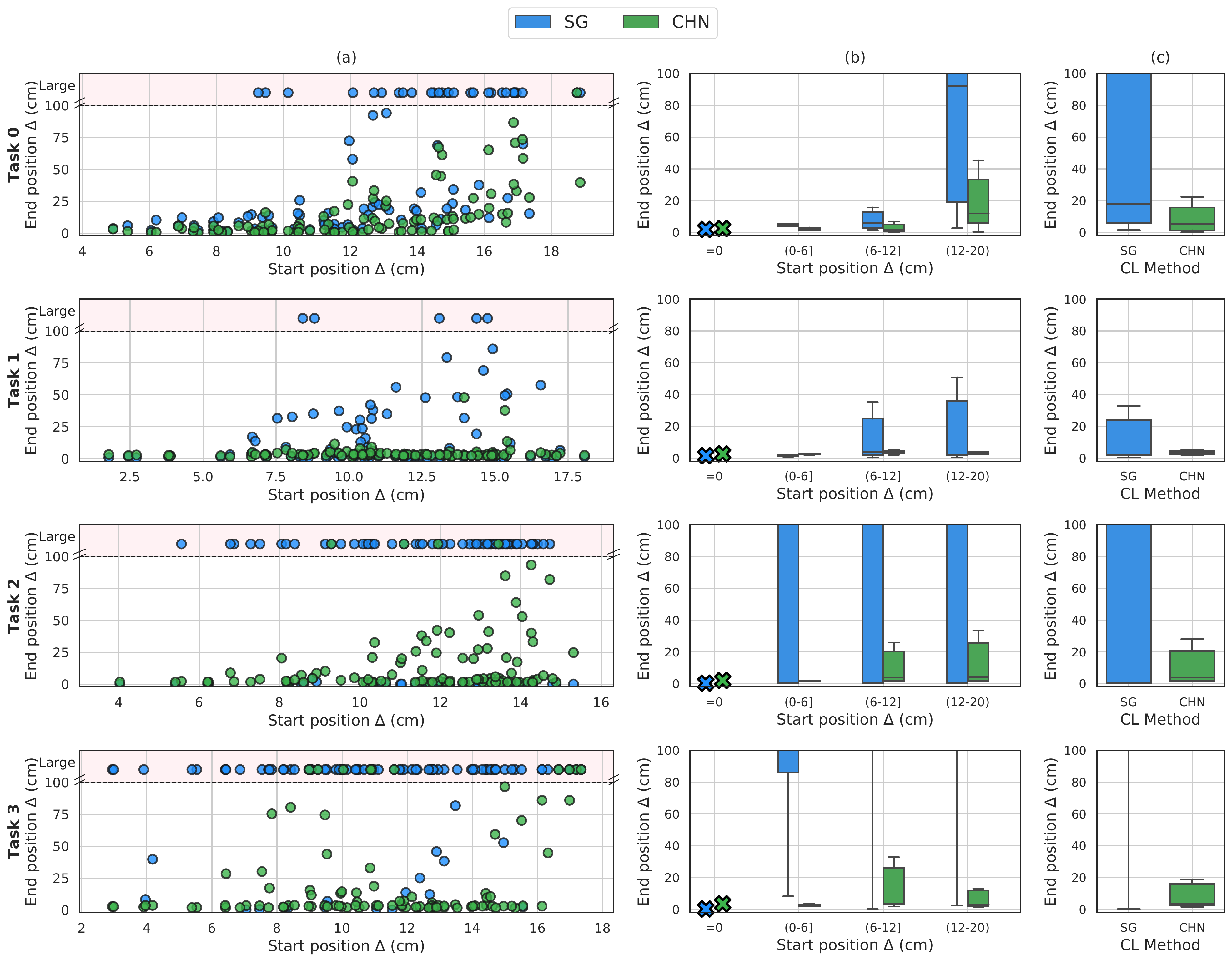}
    \caption{\change{
    Robustness of SG and CHN on the RoboTasks dataset. Each row shows the results for 1 of the 4 tasks of this dataset. Columns (a,b) show that for small values of \emph{start position $\Delta$}, SG's trajectories often diverge far from the goal position but CHN's trajectories do not. For larger values of \emph{start position $\Delta$}, CHN gets worse but much less than SG. In column (a), values greater than \emph{end position $\Delta$}=100 cm are grouped together and shown in the shaded region marked ``Large''. Column (b) shows the divergence of SG and CHN by grouping the results based on the \emph{start position $\Delta$}. Here, the symbol $ \boldsymbol{\times}$ shows the value of \emph{end position $\Delta$} when \emph{start position $\Delta$}=0. Column (c) shows overall divergence of SG and CHN for novel initial positions, where CHN shows much less divergence. All models use NODE-T for learning.
    }}
    \label{fig:robustness}
\end{figure*}


\change{

\subsection{Robustness to Starting Position}
\label{sec:appendix_robustness_start}


To test the robustness of the different CL methods to changes in the initial positions of the robot (i.e., when the robot starts at a position that is different from the start position of the demonstrated trajectory), we perform an analysis using the \mbox{RoboTasks} dataset. 

For each of the four tasks of this dataset, we randomly select one ground truth demonstration. We then create a sphere with a radius of 20 cm (the maximum reach of the Franka Emika Panda robot is 85.5 cm) centered at the starting position of the ground truth trajectory. From within the volume of this sphere, we uniformly sample 100 random starting points. We then take the trained models of SG and CHN (trained on all the tasks of the dataset), and use these to predict trajectories for each of the 4 tasks and each of the 100 random starting points with the same timesteps as the ground truth demonstration. 
By sampling points from a sphere of radius 20 cm, we are asking the models to extrapolate from the demonstrations and start from areas for which no demonstration has been provided.

For each predicted trajectory, we measure
\begin{enumerate*}[label=(\roman*)]
	\item \emph{start position $\Delta$}: the distance between the random starting position and the starting position of the demonstration;
	\item \emph{end position $\Delta$}: the distance between the end point of the predicted trajectory (that starts from a random starting position) and the end position (goal) of the demonstration.
\end{enumerate*}
The results are shown in \myfigure{fig:robustness}. In the ideal case, the end position $\Delta$ would always be 0 irrespective of the start position $\Delta$ (every trajectory, irrespective of the starting position would always converge at the goal). In the absence of any stability guarantees, if the starting positions of the robot are very different from the ground truth, the predicted trajectory can be expected to diverge far from the ground truth goal, leading to a large end position $\Delta$.

\begin{figure*}[t!]
    \centering
    \includegraphics[width=0.9\textwidth]{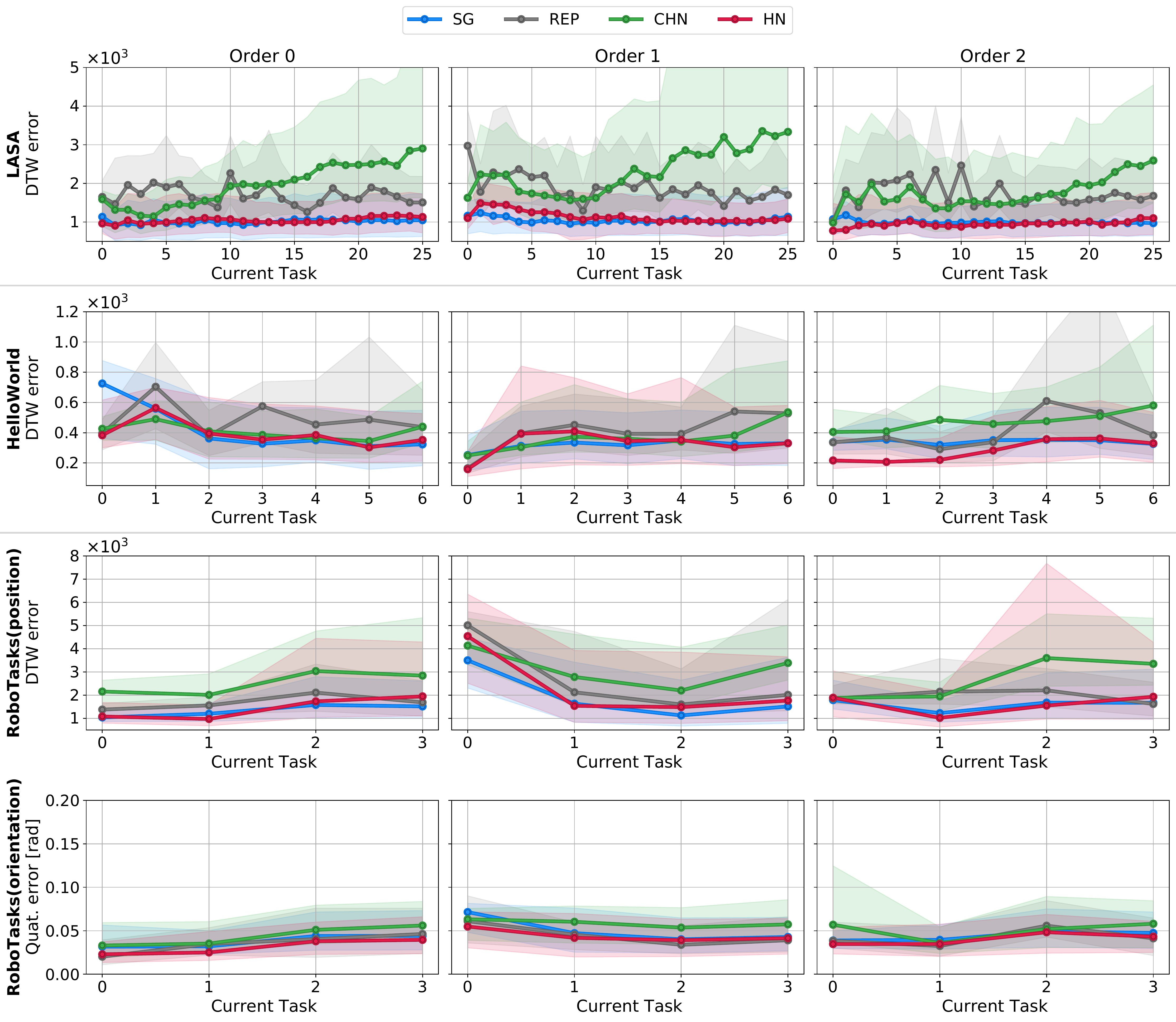}
    \caption{\change{Robustness to task order during training. The models for SG, REP, CHN and HN are trained and evaluated with 3 different task orders for each dataset (each column of plots shows the result of 1 order).
Overall the same trends are visible for all task orders. For LASA, HN is almost identical to SG, and CHN performs well for a small number of tasks but degrades afterwards. For HelloWorld, HN and CHN perform as well as the other baselines. For RoboTasks, again HN and CHN perform comparably to the other baselines. This shows that the continual learning performance is not affected by the order in which tasks are learned.
The x-axis in the plots denotes the task number. To see the name of a task from a given order, please refer to \mytable{tab:task_order} ( e.g., for LASA, task 0 from \emph{Order 0} is `Angle', while task 0 from \emph{Order 1} is `Spoon'). Also, note that the DTW metric is dependent on the dimension of the trajectory points. Thus, the DTW values for a 2D dataset such as HelloWorld are lower than those for RoboTasks. In all cases, the upper baseline SG provides a reference for comparison.
For each model, each dataset and each task order, we train and evaluate with 3 independent seeds. Similar to \myfigure{fig:lasa_cumu}, \myfigure{fig:hw_cumu} and \myfigure{fig:robottasks_cumu}, in this plot lines show median values and shaded areas represent the IQR of the errors (lower is better). 
}}
    \label{fig:task_order}
\end{figure*}

It can be seen that for all tasks, when the random starting position is up to 6 cm from the GT source, the CHN model produces trajectories that end up very close to the GT goal. This is not true for SG. Although SG performs almost perfectly when starting from the GT source position (see the $\boldsymbol{\times}$ symbols in column (b) of \myfigure{fig:robustness}, and also Figs. \ref{fig:robottasks_cumu} and \ref{fig:robottasks_pos_traj}), it produces trajectories that diverge far from the GT goal even when the random starting positions are quite close to the GT source position. For larger start position differences, the performance of CHN does degrade, but much less than that for SG, which most of the time produces trajectories that end up far beyond the reachable task space of the robot (in the scatter plots we club together all predictions that end up more than 100 cm from the goal). This shows that the continual learning process for CHN has a regularization effect which prevents the strong overfitting to demonstrations as exhibited by SG.
}

\change{

\subsection{Robustness to Task Order}
\label{sec:appendix_robustness_task_order}

\begin{table}[b!]
	\caption{Task orders. }
	\subfloat[LASA]{
		\centering
		\resizebox{0.95\textwidth}{!}{
		\begin{tabular}{rlll}
		\toprule
		 Task ID &     Order 0 file &     Order 1 file &     Order 2 file \\
		\midrule
		       0 &            Angle &            Spoon &       BendedLine \\
		       1 &       BendedLine &             Sine &           RShape \\
		       2 &           CShape &           Zshape &        Trapezoid \\
		       3 & DoubleBendedLine &           NShape &           WShape \\
		       4 &           GShape &           PShape &           PShape \\
		       5 &             heee &             heee &             heee \\
		       6 &         JShape\_2 &            Angle &           Sshape \\
		       7 &           JShape &          Khamesh &             Line \\
		       8 &          Khamesh &             Line &           JShape \\
		       9 &           Leaf\_1 &           LShape & DoubleBendedLine \\
		      10 &           Leaf\_2 &         JShape\_2 &             Sine \\
		      11 &             Line &           WShape &           GShape \\
		      12 &           LShape &           Leaf\_2 &           Sharpc \\
		      13 &           NShape &           RShape &           LShape \\
		      14 &           PShape &           JShape &           Saeghe \\
		      15 &           RShape &           Sharpc &          Khamesh \\
		      16 &           Saeghe &            Snake &           CShape \\
		      17 &           Sharpc & DoubleBendedLine &           Zshape \\
		      18 &             Sine &           Saeghe &         JShape\_2 \\
		      19 &            Snake &             Worm &            Snake \\
		      20 &            Spoon &           Leaf\_1 &            Angle \\
		      21 &           Sshape &           GShape &             Worm \\
		      22 &        Trapezoid &       BendedLine &           Leaf\_2 \\
		      23 &             Worm &           Sshape &            Spoon \\
		      24 &           WShape &           CShape &           NShape \\
		      25 &           Zshape &        Trapezoid &           Leaf\_1 \\
		\bottomrule
		\end{tabular}
		}
		\label{tab:task_order_lasa}
	 	}	
	 	
	\subfloat[HelloWorld]{
		\centering
		\resizebox{0.7\textwidth}{!}{
		\begin{tabular}{rlll}
		\toprule
		 Task ID & Order 0 file & Order 1 file & Order 2 file \\
		\midrule
		       0 &            h &            l &            o \\
		       1 &            e &            d &            w \\
		       2 &            l &            h &            l \\
		       3 &            o &            o &            h \\
		       4 &            w &            w &            e \\
		       5 &            r &            r &            d \\
		       6 &            d &            e &            r \\
		\bottomrule
		\end{tabular}
		}
		\label{tab:task_order_hw}
	 	}	
	 	
	\subfloat[RoboTasks]{
		\centering
		\resizebox{0.99\textwidth}{!}{
		\begin{tabular}{rlll}
		\toprule
		 Task ID &         Order 0 file &         Order 1 file &         Order 2 file \\
		\midrule
		       0 &              openbox & platestandingtolying &              pouring \\
		       1 &         bottle2shelf &              openbox &              openbox \\
		       2 & platestandingtolying &         bottle2shelf & platestandingtolying \\
		       3 &              pouring &              pouring &         bottle2shelf \\
		\bottomrule
		\end{tabular}
		}
		\label{tab:task_order_rt}
	 	}	
	\label{tab:task_order}
\end{table}

We test the robustness of the best performing models (SG, REP, CHN and HN) against changes in the order in which tasks are presented during training. 
For each of the 3 datasets used in this paper, we create 3 different orders of tasks: \emph{Order 0}, \emph{Order 1}, and \emph{Order 2}. The first order (Order 0) is the same as that used for the results reported in \mysection{sec:results}. The next 2 task orders for each dataset are created by randomly shuffling the first task order of the respective datasets. The task orders and the corresponding data file names used during training are listed in \mytable{tab:task_order}. 

We train and evaluate models for SG, REP, CHN, and HN (as they perform the best) each with 3 independent seeds on each task order of each dataset. The results for this evaluation are shown in \myfigure{fig:task_order}. 
Although the results are marginally different for the different task orders, the overall trends are similar across task orders for the LASA, HelloWorld and RoboTasks datasets. 

For LASA (see the first row of plots in \myfigure{fig:task_order}), the performance of HN is almost identical to that of the upper baseline SG, while REP and CHN perform worse for all orders. CHN performs well up to a small number of tasks but its performance degrades as more tasks are learned.

The results for HelloWorld (second row of plots in \myfigure{fig:task_order}) also show similar trends for all 3 task orders: CHN performs similarly to SG, REP and HN, in spite of being much smaller in size than SG and HN, and without needing to store training data of past tasks like REP.

Similarly, for RoboTasks (last two rows of plots in \myfigure{fig:task_order}), the task order does not seem to affect the continual learning performance. Learning a difficult task is equally challenging for all the methods (e.g. the \emph{plateslyingtostanding} task, which is task 2 in Order 0 and task 0 in Order 1, appears to be a difficult task judging from the relatively higher errors). However this does not affect the relative performance of the different methods. CHN performs almost equally well as the other methods, although other methods have more parameters (SG, HN) or store past training data (REP).
}

\end{document}